\documentclass[12pt]{article}

\usepackage{amsmath,graphicx}
\DeclareMathOperator*{\argmin}{arg\,min}
\DeclareMathOperator*{\argmax}{arg\,max}
\usepackage{algorithm}
\usepackage[noend]{algpseudocode}
\usepackage{mathtools}
\usepackage{url}
\usepackage{makecell}
\usepackage{subcaption}

\begin{document}
\title{Detection of Backdoors in Trained Classifiers Without Access to the Training Set}

\author{Zhen Xiang, David J. Miller, and George Kesidis
	\thanks{The authors are with the School of Electrical Engineering and Computer
		Science, The Pennsylvania State University, University Park, PA, 16802, USA 
		(\{zux49,djm25,gik2\}@psu.edu). 
		The last two authors are also with Anomalee Inc.}
}

\maketitle

\begin{abstract}

With wide deployment of deep neural network (DNN) classifiers, there is great potential for harm from adversarial learning attacks. Recently, a special type of data poisoning (DP) attack, known as a backdoor, was proposed. These attacks do not seek to degrade classification accuracy, but rather to have the classifier learn to classify to a target class $t^{\ast}$ whenever the backdoor pattern is present in a test example originally from a source class $s^{\ast}$. Launching backdoor attacks does not require knowledge of the classifier or its training process -- only the ability to poison the training set with exemplars containing a backdoor pattern (labeled with the target class). Defenses against backdoors can be deployed before/during training, post-training, or at test time. Here, we address post-training detection in DNN image classifiers, seldom considered in existing works, wherein {\it the defender does not have access to the poisoned training set}, but only to the trained classifier itself, as well as to clean (unpoisoned) examples from the classification domain. This scenario is of great interest because {\it e.g.} a classifier may be the basis of a phone app that will be shared with many users. Detection may thus reveal a widespread attack. We propose a purely unsupervised anomaly detection (AD) defense against imperceptible backdoor attacks that: i) detects whether the trained DNN has been backdoor-attacked; ii) infers the source and target classes in a detected attack; iii) estimates the backdoor pattern itself. Our AD approach involves learning (via suitable cost function minimization) the minimum size/norm perturbation (putative backdoor) required to induce the classifier to misclassify (most) examples from class $s$ to class $t$, for all $(s,t)$ pairs. Our hypothesis is that non-attacked pairs require large perturbations, while the attacked pair $(s^{\ast}, t^{\ast})$ requires much smaller ones. This is convincingly borne out experimentally. We identify a variety of plausible cost functions and devise a novel, robust hypothesis testing approach to perform detection inference.  We test our approach, in comparison with the state-of-the-art methods, for several backdoor patterns, attack settings and mechanisms, and data sets and demonstrate its favorability. Our defense essentially requires setting a single hyperparameter (the detection threshold), which can {\it e.g.} be chosen to fix the system's false positive rate.
\end{abstract}

\vspace{-0.15in}
\section{Introduction}\label{sec:introduction}
As deep neural network (DNN) classifiers are increasingly used in many applications, including security-sensitive ones, they are becoming valuable targets for adversaries who intend to ``break'' them. Over the past decade, {\it adversarial learning} research in devising attacks against classifiers and defenses against such attacks has quickly developed \cite{Review}. From a broad view, there are several prominent types of attacks. Test-time evasion (TTE) attacks induce misclassifications during operation/test-time by modifying test samples in a human-imperceptible (or machine-evasive) fashion, {\it e.g.} \cite{Szegedy_seminal}\cite{Papernot}\cite{Goodfellow}\cite{DeepFool}\cite{Yuan2019}. Data poisoning (DP) attacks insert malicious samples into the training set, {\it often} to degrade the classifier's accuracy \cite{Tygar11}\cite{Xiao15}. Reverse engineering (RE) attacks learn the decision rule of a black-box classifier through numerous queries to the classifier \cite{Reiter}\cite{Papernot2}\cite{ICASSP19}.

\vspace{-0.1in}
\subsection{Backdoor DP Attacks}

Recently, a new form of DP attack -- a {\it{backdoor}} attack -- was proposed, usually\footnote{Backdoor attacks could also be launched against a speech recognizer \cite{Trojan} or text classifier \cite{LSTMBD}.} against DNN {\it image} classifiers \cite{Song}\cite{BadNet}\cite{Trojan}\cite{Haoti}. Under such attacks, a relatively small number of legitimate examples from one or more source classes, but containing the same embedded backdoor pattern and (mis)labeled to a target class, are added to the training set\footnote{The source class(es), the target class and the backdoor pattern are all decided by the attacker.}. Then during testing, the classifier trained on this poisoned training set will likely (mis)classify examples from those source classes to the target class if the same backdoor pattern is present. The backdoor pattern, principally designed to be evasive to possible occasional human inspection of the training set, could be: 1) a human-imperceptible perturbation to clean images \cite{SS}\cite{Haoti}\cite{CI}; 2) a seemingly innocuous visible object in a scene, e.g. a bird flying in the sky \cite{Song}\cite{BadNet}\cite{Perceptible_MLSP}. 
Moreover, the perturbation could be applied additively, multiplicatively, or through some other operation.  While we focus on additive perturbations, we will demonstrate our approach is also effective for other perturbative mechanisms, without requiring knowledge of the mechanism used.
What makes backdoor attacks attractive is that successful attacks will {\it not} degrade the classifier's accuracy on clean examples. 
Thus, validation/test set accuracy degradation that could be observed for a {\it classical} DP attack cannot be reliably used as a basis for detecting backdoor attacks. Moreover, launching backdoor DP attacks only requires the ability to poison the training set (with a relatively modest number of poisoned training samples) -- {\it no} knowledge of the classifier (or some trained surrogate classifier with similar classification performance) is required, unlike for a TTE attack.
The attacker's poisoning capability is facilitated by the need in practice to obtain ``big data'' suitable for accurately training a DNN for a given domain -- to do so, one may
need to seek data from as many sources as possible (some of which could be
attackers).
Since backdoor attacks require less attacker knowledge and capabilities than {\it e.g.} TTE attacks, they are a serious practical threat to the integrity of deployed machine learning solutions. Note also that detection defenses that have been developed for TTE attacks, e.g. \cite{ADA-NC}, do not have obvious applicability for detecting backdoor attacks -- these two attacks are very different.

\vspace{-0.1in}
\subsection{Three Defense Scenarios}

While backdoor attacks on DNN classifiers appeared only fairly recently, several defenses have been proposed aiming to detect and mitigate them \cite{SS}\cite{AC}\cite{FP}\cite{CI}\cite{NC}\cite{Perceptible_MLSP}. We identify three defense scenarios. The first is before/during-training, where the defender has access to the (possibly poisoned) training set and either to the training process or to the trained (attacked) classifier. The defender's goal in this setting is to detect whether the training set has been poisoned by a backdoor or not. If the training set was poisoned, the malicious examples should be identified and removed before classifier training (or retraining)\footnote{It is additionally useful to determine the source and target classes involved in the detected backdoor.}. Recently proposed backdoor defenses (e.g. \cite{SS}\cite{AC}\cite{CI}) for this case first train a network using the potentially poisoned training set and then investigate the distribution of feature vectors produced {\it e.g.} at the penultimate layer of the DNN, for each class. If a target class has been attacked, the penultimate layer feature vectors for the backdoor patterns should be distinct (separable) from the clean samples from this class. Among existing works, a state-of-the-art defense \cite{CI} first clusters the feature vectors from each class into a Bayesian Information Criterion (BIC)-selected \cite{Schwarz} number of mixture components; then the components are separately evaluated (for possible removal from the training set) according to a ``cluster impurity'' measure.

The second defense scenario is ``in-flight'', where backdoor detection is performed during the classifier's use/operation, when it is presented with a test example. The defender's goal is to detect potential backdoor patterns,
possibly identify the entity making use of the backdoor,
 and trigger an alarm \cite{Review}. To our knowledge, no existing works have devised defenses against backdoors for this scenario.

\begin{figure}[h]	
	\begin{minipage}{1.0\linewidth}
		\centering
		\centerline{\includegraphics[width=7cm]{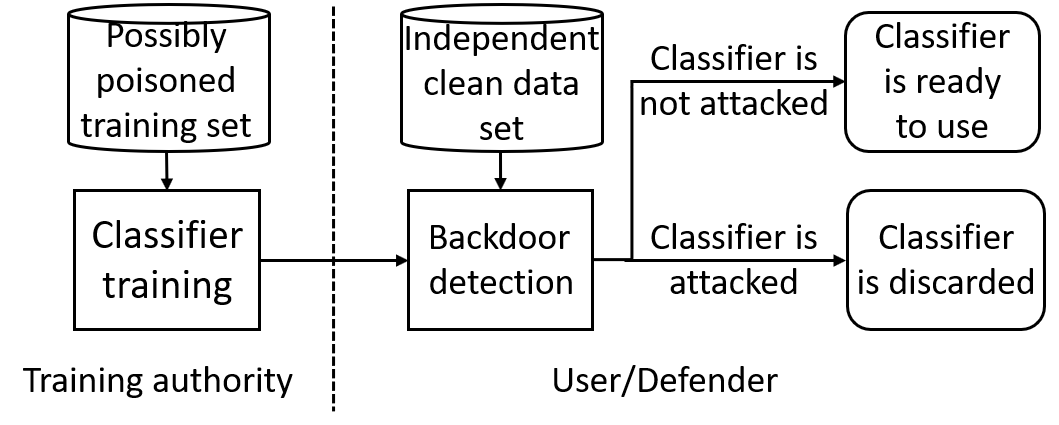}}
	\end{minipage}
	\caption{Illustration of the post-training scenario. The user/defender receives a classifier trained on a possibly (backdoor) poisoned training set from the training authority. An independent clean data set is used to detect whether the classifier is backdoor attacked. If a detection is made, the classifier is discarded; else, it is certified for use.}
	\label{fig:post_training}
	\vspace{-0.15in}
\end{figure}

This paper focuses on the third scenario, {\it{post-training}}, where the defender has access to the trained DNN but {\it not} to the possibly backdoor-poisoned training set used for its learning. We also assume that an independent clean data set is available (no backdoors present) with examples from each of the classes from the domain \cite{FP}\cite{NC}\cite{Tabor}. Note that this independent set is {\it not} the clean training set -- it is whatever labeled examples from the domain that the defender has access to. Availability of such data in practice is reasonable -- it is not a strong assumption\footnote{For example, there may be two databases available for dog breed classification. If database A is used for classifier learning, database B can be used as the independent data set by the defender.}. The post-training scenario is of strong practical interest because DNN training can be extremely computationally intensive; hence learning may be outsourced to a third party who may be compromised \cite{BadNet}. Moreover, there are many pure {\it consumers} of machine learning systems -- {\it e.g.}, an app used on millions of cell phones. The app user will not have access to the training set on which the app's classifier was learned. Still, the user would like to know whether the app's classifier has been backdoor-poisoned. The user may reasonably possess clean labeled data on which, {\it e.g.}, it can evaluate the performance of the DNN. Consider also legacy classifier systems, for which the training data may be lost or inaccessible. Since the attacker seeks for the backdoor mapping to be learned {\it{without}} affecting the classifier’s accuracy on clean (backdoor-free) samples, such attacks may be inherently evasive. As the user/defender, our fundamental goal is to detect whether the DNN has been backdoor-attacked or not, using only the independent clean data set. If attacked, the DNN should be discarded and replaced by a new one from a more trustworthy source. This is illustrated in Figure \ref{fig:post_training}. Beyond that, we also aim to infer the source class(es) and the target class of the detected backdoor attack. Additionally, we aim to estimate the backdoor pattern itself, given that an attack has been detected. The backdoor could be mitigated if the DNN is fine-tuned using a (sufficient number of) clean examples with the estimated backdoor pattern added in (and labeled with the {\it correct} source class -- not the backdoor's target class). Such tuning may ``unlearn'' the backdoor. If such tuning is not practical, the estimated backdoor pattern may still help to detect and thus reject use of the backdoor operationally, {\it i.e.} ``in-flight''. This may be considered in our future work.

\vspace{-0.1in}
\subsection{Proposed Anomaly Detection Framework}\label{subsec:intro_AD}

In this paper, we propose a {\it purely unsupervised} anomaly detection (AD) approach thwarting backdoor attacks post-training. Note that there is sufficient interest in post-training backdoor detection that it is the subject of a recent IARPA BAA \cite{BAA}. While this BAA allowed detector solutions to make use of supervised exemplar DNN classifiers (``with backdoor present'', and ``without backdoor''), such labeled information is unrealistic in practice and is not needed by our AD approach. Here we mainly focus on detecting attacks with imperceptible backdoor patterns that are additive perturbations -- {\it imperceptible} effectively means that the perturbations are {\it small}, as measured by some norm metric. Small perturbations are invisible to humans and at the same time not so readily machine-detected. We will also demonstrate that imperceptible backdoors with other perturbative mechanisms can be effectively detected by a slight generalization of our AD approach.
Post-training detection for {\it perceptible} backdoor patterns, i.e. visible objects innocuous to the scene, is studied in \cite{Perceptible_MLSP}. 
Our detector, though targeting backdoor DP attacks, is inspired by an observation made about TTE attacks: {\it for a DNN trained on a clean training set, for any $(s, t)\in\mathcal{C}\times\mathcal{C}$ (source, target) class pair, with $\mathcal{C}$ the class set, altering the class decision for {\it every} image from class $s$ to class $t$ using the {\it same} additive perturbation requires much larger perturbation size/norm than for altering a single image's decision from class $s$ to class $t$} \cite{Universal}. However, if there is a class pair $(s^{\ast}, t^{\ast})$ involved in devising a backdoor attack with an imperceptible backdoor pattern, there should be a perturbation that is {\it very modest} in size/norm and yet which when added to all examples from class $s^{\ast}$ induces the DNN to change the decision to class $t^{\ast}$ for most of them -- {\it the backdoor pattern\footnote{As shown in Section \ref{sec:size_of_gt_bd}, when the backdoor pattern has large perturbation size/norm but is still human-imperceptible, there exists a perturbation with much smaller size/norm containing sufficient features of the true backdoor pattern (e.g. similar pattern but with smaller size/norm) that could still induce high group misclassification for the (source, target) class pair prescribed by the attacker.} itself is one such perturbation}. Thus, our AD defense is devised by seeking for each $(s, t)$ class pair the minimal-size/norm perturbation that induces a significantly high fraction (no less than some $\pi \in (0, 1]$) of group misclassifications from class $s$ to class $t$. 
This is captured mathematically by the following constrained optimization problem:
\begin{equation*} \tag{1}\label{eq:opt_raw}
\vspace{-0.1in}
\begin{aligned}
& \underset{\underline{v}}{\text{minimize}}
& & d(\underline{v}) \\
\vspace{-0.05in}
& \text{subject to}
& & \frac{1}{|\mathcal{D}_s|}\sum_{\underline{x}\in\mathcal{D}_s}{\bf 1} (f([\underline{x}+\underline{v}]_c)=t) ~\geq~ \pi,
\vspace{-0.05in}
\end{aligned}
\end{equation*}
where $d(\cdot)$ is the metric for measuring the size (L1 norm) or the energy (squared L2 norm) of a perturbation $\underline{v}$, $[\cdot]_c$
represents a clipping operation\footnote{The clipping operation is an alternative to imposing strict box constraints on the backdoor pattern \cite{Szegedy_seminal}.} to ensure pixel intensities stay in the allowed range, $f(\cdot)$ is the classifier's decision, and $\mathcal{D}_s$ is the set of clean patterns from class $s$. Also, ${\bf 1}(\cdot)$ is an indicator function whose value is 1 when the enclosed event occurs, and 0 otherwise.

This optimization effectively aims to reverse-engineer a possible backdoor pattern. The size/norm of the obtained perturbations for {\it all} class pairs then go through our hypothesis-test-based detection inference and a resulting order statistic p-value is derived to make detections with any user-preferred confidence level. Compared with other related works that will be reviewed in Section \ref{sec:related_work}, our defense does not make assumptions about the number of source classes involved in the potential backdoor attack. Also, we do not constrain the shape, location, or spatial support of the backdoor pattern, unlike existing approaches. We will show the effectiveness of our defense for different backdoor patterns, data sets (in the supplementary material), and attack settings, through large-scale experiments in Section \ref{sec:experiments}. We will also discuss several variants of the basic objective function we use for learning the perturbations, our choice of decision statistics and detection inference strategy, design choices, and computational complexity.

\vspace{-0.1in}
\subsection{Summary of Our Contributions}

1) We develop one of the first methods (and an approach superior to \cite{NC} and \cite{FP}, as shown in Section \ref{sec:eval_main} and the supplementary material, respectively) for detecting backdoor attacks on DNNs {\it{post-training}}, i.e. without the benefit of the poisoned training set, and also without the benefit of additional unrealistic information such as supervised labeled examples of DNNs with and without backdoors \cite{BAA}.

2) Our AD {\it detection} approach is related to TTE attack approaches, except in two key respects: i) Instead of inducing misclassification of a {\it single} image, we seek a (small) perturbation that induces misclassification of an entire {\it group} (class, or subset of classes) to a target class; ii) the gradient optimization based search for such perturbations is performed as part of an AD {\it defense}, rather than as part of an attack.

3) We formulate a novel, principled, robust detection inference approach based on the perturbation size/norm statistics produced for all $(s, t)\in\mathcal{C}\times\mathcal{C}$ pair hypotheses. Moreover, we extend our approach to account for class confusion information -- this allows us to reliably distinguish backdoor attacks from class pairs that merely possess naturally high class confusion.

4) We perform extensive experiments comparing against alternative defenses on several image databases, under different attack scenarios (single pairs, multiple source classes and one target class), and for different backdoor patterns ranging from a single pixel mask to a global image mask. Our approach is found to be highly effective at detecting attacks, and superior to other methods, across these experiments. 
We further elucidate reasons why our approach achieves superior results to existing methods.
We also make some novel experimental observations about DNN ``generalization'' induced by backdoors (a ``collateral damage'' phenomenon).

5) We give a proof of concept that our approach can be made robust to the mechanism for introducing backdoor patterns and need not assume, {\it e.g.}, that the backdoor perturbation is additive. Our approach is also effective if the unknown attack is multiplicative or involves local patch replacement \cite{NC}.

\vspace{-0.1in}
\section{Threat Model}\label{sec:threat_model}

\vspace{-0.1in}
\subsection{Notation}\label{subsec:notations}

Denote the DNN classifier to be examined for backdoor attacks as $f(\cdot): {\mathcal{X}} \rightarrow {\mathcal{C}}$, where ${\mathcal{X}}$ is the input (image) space and ${\mathcal{C}} = \{c_1, \ldots, c_K\}$ is the set of class labels. The classifier uses a ``winner take all'' rule. For example, for the commonly used softmax activation function in the output layer of DNN classifiers, the predicted label of an image $\underline{x} \in \mathcal{X}$ is
\begin{equation*}
\vspace{-0.05in}
f(\underline{x}) = \argmax_{c\in\mathcal{C}} p(c|\underline{x}).
\end{equation*}
where $p(c|\underline{x})$ is the {\it a posteriori} probability of class $c\in\mathcal{C}$. We also denote the clean labeled set used for detection as $\mathcal{D}$, which has a partition $\mathcal{D} = \cup_{c\in\mathcal{C}} \mathcal{D}_{c}$, where $\mathcal{D}_{c}$ contains all images labeled by $c${\footnote{If $\mathcal{D}$ is unlabeled, we can instead define $\mathcal{D}_{c}$ as the set of images in $\mathcal{D}$ {\it{classified by the DNN}} to $c$. Assuming the classifier's error rate for class $c$ is low, this set will be a good surrogate for the images truly from class $c$.}}.

\vspace{-0.15in}
\subsection{Background on Backdoor Attacks}\label{subsec:backgrouds}

A backdoor attack is launched with a (set of) source class(es) $s^{\ast}\in\mathcal{C}$ ($\mathcal{S}^{\ast}\subset\mathcal{C}$ for multiple source classes), a target class $t^{\ast}\in\mathcal{C}$ with $t^{\ast}\neq s^{\ast}$ ($t^{\ast}\notin\mathcal{S}^{\ast}$ for multiple source classes), and a backdoor pattern $\underline{v}^{\ast}$, all prescribed by the attacker. Like most related works, we assume a single target class and a single backdoor pattern for each attack for simplicity \cite{NC, Tabor, SS, CI}. But we allow backdoor attacks to have {\it any} possible number of source classes (ranging from $1$ up to $(K-1)$, for $|{\mathcal C}|=K$ legitimate classes). Although more source classes may make the attack less evasive in practice, we investigate the multiple source class case to demonstrate the detection capability of our AD framework under general attack settings.

In this work, we consider an imperceptible backdoor pattern $\underline{v}^{\ast}$ with the same dimensionality as the input image. Practical backdoor patterns can also be human-perceptible but seemingly innocuous objects (e.g. a bird in the sky, a pair of glasses on a face \cite{Song}); however, the scene-plausibility of such backdoors is highly dependent on the image domain.  Such backdoor attacks may also require much more crafting effort\footnote{Large crafting effort would not be required if the backdoor involves simply putting glasses on a face -- however, such a backdoor might also be rejected by the security protocol of a face recognition system (no glasses allowed).  Moreover, one might discover the backdoor by accident, simply by (innocently) putting on a pair of glasses and noticing the change in the classifier's decision.}. The defense developed in this paper is not intended to detect backdoors based on innocuous, {\it perceptible} patterns -- detection of such backdoors is addressed in \cite{Perceptible_MLSP}.

An imperceptible backdoor pattern can be embedded into a clean image $\underline{x}$ as an additive perturbation (\cite{SS, CI, Haoti, Hidden-trigger}), by
\begin{equation*}\tag{2}\label{eq:bd_embedding_imperceptible}
\vspace{-0.05in}
B_{\rm add}(\underline{x}, \underline{v}^{\ast}) = [\underline{x}+\underline{v}^{\ast}]_c,
\end{equation*}
where the clipping operation $[\cdot]_c$ constrains the pixel intensity values to their valid range ({\it e.g.} $[0,255]$ for a gray scale intensity). Backdoor patterns shown to have good imperceptibility and attack effectiveness include ``global'' patterns with large $||\underline{v}^{\ast}||_0$ but very small $||\underline{v}^{\ast}||_2$ \cite{Haoti, Hidden-trigger, Song}, and ``local'' patterns with small $||\underline{v}^{\ast}||_0$ (i.e. perturbing only few pixels) and modest $||\underline{v}^{\ast}||_2$ \cite{SS, CI}.

Our AD framework {\it does not} make any assumptions about the shape, location, or spatial support of the backdoor pattern -- it detects attacks with either global or local imperceptible backdoor patterns, as shown by our experiments. Moreover, while we mainly focus on additive perturbations (embedded by Eq. (\ref{eq:bd_embedding_imperceptible})) in our experiments, we also show that our generalized AD framework can effectively detect attacks with other perturbative mechanisms (see Section \ref{subsubsec:bd_multi}). Finally, we do not consider finite precision effects in this work, associated with very low bit rate (compressed) representation of images, {\it i.e.} we allow $v_{ij}^{\ast}$, the perturbation for pixel $(i,j)$, to be any real number in the valid range and likewise the clipped version of $x_{ij} + v_{ij}^{\ast}$ can be any real number in the valid range.

\vspace{-0.15in}
\subsection{Attacker's Goals, Knowledge, and Assumptions}\label{subsec:attacker_assump}
The goal of backdoor DP attackers is to induce the classifier to learn to misclassify samples that ground-truth come from the source class $s^{\ast}$ (or any class $c\in\mathcal{S}^{\ast}$) to the target class $t^{\ast}$ whenever the attacker's chosen backdoor pattern $\underline{v}^{\ast}$ is embedded to an original image from class $s^{\ast}$. This is achieved by poisoning the classifier's training set with examples from class $s^{\ast}$ that are ({\it e.g.} additively) perturbed with the backdoor pattern and mislabeled as originating from class $t^{\ast}$.  
The attack is evaluated by the attack success rate on a test set of patterns from class $s^{\ast}$ with the backdoor pattern $\underline{v}^{\ast}$ added to each such test example. Also, the test accuracy on clean (backdoor-free) samples should not be degraded. In addition, the devised attack should be {\it evasive}, such that the backdoor patterns are both difficult to machine-detect and difficult for a human to perceive (in case there is human training set inspection). 
The knowledge and capabilities of backdoor attackers that are assessed in this work are as follows:

1) The attacker has full knowledge of the legitimate image domain and access to the training set. Thus, the attack can be launched by simply adding a backdoor pattern to relatively few legitimate images from $s^{\ast}$ (or several classes not including $t^{\ast}$), labeling them as coming from  $t^{\ast}$, and inserting them into the training set. At test time, the attacker can then add the backdoor pattern to an image from $s^{\ast}$ and, with high probability, elicit a classifier decision to $t^{\ast}$. Although such attacks do not truly require knowledge about the classifier or the training process, we will endow the attacker with this knowledge. In particular, with this knowledge the attacker can replicate the training process using different candidate backdoor patterns. The one with lowest perturbation size/norm{\footnote{Perturbations with small size/norm are not only imperceptible to humans. In \cite{Review} and \cite{CI}, such backdoor patterns are experimentally shown to be evasive to several existing backdoor defenses (before/during-training) that are very successful against {\it human-perceptible} backdoor patterns \cite{AC}\cite{SS}.}} among those with an acceptable attack success rate and negligible degradation in test accuracy on clean samples is then chosen for the actual backdoor attack -- in this way,
the attacker seeks the least human-perceptible attack (using perturbation size/energy as a surrogate
for perceptibility).

2) The attacker has no specific knowledge about any defenses that may be deployed to detect the attack. 
However, the attacker does know that the backdoor pattern should be human-imperceptible.

\vspace{-0.125in}
\subsection{Defender's Goals, Knowledge and Assumptions}\label{subsec:defender_assump}

In this paper, we play the role of a defender who aims to detect backdoors in trained networks with low false positives and to infer the true source class(es) and target class. An additional goal is to estimate the backdoor pattern, which can possibly be used to mitigate the attack as discussed in Section \ref{sec:introduction}. The capabilities possessed by the defender are:

1) The defender possesses an independent labeled, clean data set with images from all classes (i.e. $\mathcal{D}$), but does not have access to the (possibly poisoned) training set used to design the classifier. This unavailability of the training set is what distinguishes post-training detection from before/during-training detection. 
This also makes post-training detection both an interesting and challenging problem, as any information about a possible backdoor is only {\it latently} available in the learned DNN weights.

2) The defender has full access to the classifier, but no training/retraining (using the available clean data set) is allowed.  Such retraining on clean data would of course remove the learned backdoor.  However, the clean data set available to the user will generally be too small to adequately retrain the classifier and/or the user may have inadequate computational resources for such retraining.

3) The defender assumes only that the attack, if present, involves a single target class and that the attacker will seek to make the backdoor human-imperceptible. No knowledge of the backdoor pattern, the number of source classes, the label(s) of the source class(es) or the label of the target class are available to the defender.  For much of this work, the backdoor pattern is additively applied and this is known to the defender.  However, we will also show an effective detector can be devised for more general attack mechanisms ({\it e.g.}, multiplicative perturbations), and without requiring knowledge of the perturbative mechanism used by the attacker.
\vspace{-0.15in}
\section{Detection Methodology}\label{sec:detection}


\vspace{-0.05in}
\subsection{Proposed Detector: Reverse-Engineering the Attack}\label{subsec:detection}

\subsubsection{Key Idea}
As noted earlier, {\it the premise behind the proposed AD framework is that for a classifier that has been backdoor data poisoned
with source class $s^{\ast}$ and target class $t^{\ast}$, the required perturbation size for a common perturbation to 
induce misclassification to $t^{\ast}$ for most images from class $s^{\ast}$ is {\it  much smaller} than for class pairs that have not been backdoor-poisoned -- in fact, one such common perturbation (it need not be unique) is
the backdoor pattern $\underline{v}^{\ast}$ itself.}  Thus, if one can find a small perturbation that
induces most patterns from $s$ to be misclassified to $t$ this is indicative that the
DNN is the victim of an (imperceptible) backdoor attack involving the class pair $(s,t)$.

In our detection scenario, for each class pair $(s, t)\in\mathcal{C}\times\mathcal{C}$, we estimate the optimal perturbation $\underline{v}_{st}^{\ast}$ that induces at least $\pi$-level group misclassification as the solution to (\ref{eq:opt_raw}) in Section \ref{subsec:intro_AD}.  $\pi \in (0, 1]$ can be considered the minimum misclassification rate to deem a backdoor attack ``successful'' -- {\it e.g.}
we might set $\pi=0.8$.
The choice of $\pi$ does not have strong effect on our detection performance
because the required perturbation size for a true attack pair $(s^{\ast},t^{\ast})$ is observed to be anomalously small, compared with the size for non-attack class pairs, over a large range of $\pi$ values, as will be shown in Section \ref{sec:choice_of_pi}. In practice $\pi$ can be set by the user.

Our detection procedure consists of two steps. First, we estimate $\underline{v}_{st}^{\ast}$ for each class pair $(s, t)\in\mathcal{C}\times\mathcal{C}, s \neq t$. Second, an inference procedure is performed based on the collection of statistics $\{d(\underline{v}_{st}^{\ast})\}$ for $\forall (s,t)$. If the classifier is attacked and the backdoor involves, for example, a single (source, target) class pair $(s^{\ast}, t^{\ast})$, we would expect $d(\underline{v}_{s^{\ast}t^{\ast}}^{\ast}) \ll d(\underline{v}_{st}^{\ast})$ for all $(s, t) \neq (s^{\ast}, t^{\ast})$. If the network has not been attacked, we would expect there are no such anomalies. Our actual inference procedure is detailed in Section \ref{sec:inference}.

\subsubsection{Perturbation Optimization}

Unfortunately, (\ref{eq:opt_raw}) does not have a closed-form solution and cannot be solved using gradient-based methods because of the non-differentiable indicator function. Hence, instead we choose
the perturbation to minimize a {\it surrogate} objective function. Note that choosing a perturbation to induce group {\it misclassification}  is quite reminiscent of optimization of a parameterized classification model to maximize the {\it correct} classification rate.  Further, note that there is no singular objective function used in practice for learning good classifiers -- cross entropy, discriminative learning that seeks to minimize a soft error count measure \cite{Juang}, classifier margin, as well as other training objectives \cite{Duda}
have all been shown to be ``good'' surrogate objectives for the (non-differentiable) classification error rate in that minimizing them to determine the classifier
leads to good (test set) classification accuracy in practice. Similarly, there are multiple plausible surrogate objectives for (\ref{eq:opt_raw}).  While in Section \ref{sec:surrogate} we discuss various alternatives, in our experiments
we have primarily optimized the perturbation ({\it i.e.} estimated putative backdoors) by performing gradient descent on the objective function
\begin{equation*} \tag{3}\label{eq:obj_basic}
\vspace{-0.05in}
J_{st}(\underline{v}) = -\frac{1}{|\mathcal{D}_s|}\sum_{\underline{x}\in\mathcal{D}_s} p(t|[\underline{x}+ \underline{v}]_c),
\vspace{-0.025in}
\end{equation*}
until the constraint in problem (\ref{eq:opt_raw}) is satisfied.
The algorithm is summarized below.

\begin{algorithm}
	\vspace{-0.05in}
	\caption{Perturbation Optimization}\label{alg:main}
	\begin{algorithmic}[1]
		\State Initialization: $\underline{v} \gets \underline{0}$, $\rho \gets \frac{1}{|\mathcal{D}_s|}\sum_{\underline{x}\in\mathcal{D}_s}{\bf 1} (f(\underline{x})=t)$
		\While {$\rho < \pi$}
		\State $\underline{v} \gets \underline{v} - \delta\cdot\nabla J_{st}(\underline{v})$
		\State $\rho \gets \frac{1}{|\mathcal{D}_s|}\sum_{\underline{x}\in\mathcal{D}_s}{\bf 1} (f([\underline{x}+\underline{v}]_c)=t)$
		\EndWhile
	\end{algorithmic}
\end{algorithm}
\vspace{-0.1in}

There are several things to note about Algorithm \ref{alg:main}. First, $\rho$ is updated in each iteration as the misclassification fraction induced by the current perturbation $\underline{v}$. 
Second, note that we do not explicitly impose a constraint on the perturbation size -- the perturbation is initially set to zero, with its size tending to grow with iterations.  While a smaller-sized perturbation inducing $\pi$-level misclassification could be achieved by minimizing $J_{st}(\cdot)$ subject to an explicit constraint on the perturbation size\footnote{Or by minimizing the perturbation size subject to a level of $J_{st}(\cdot)$.}, this would also lead to an optimization problem that would require the choice of a Lagrange multiplier specifying the value of the constraint -- appropriate (likely search-entailed) choice of the Lagrange multiplier would complicate this constrained optimization problem.  In practice, we have found that gradient descent on $J_{st}(\cdot)$ with termination once $\pi$-level misclassification is achieved
yields small perturbations for backdoor pairs {\it relative} to those required for non-backdoor pairs. 
This is sufficient for successful anomaly detection of backdoor pairs.
Third, we recognize that the gradient of $J_{st}(\cdot)$ has non-zero contributions from samples even once they
are successfully misclassified.  Again, we do not claim that $J_{st}(\cdot)$ is the best surrogate for misclassification count that could be used.  While we have found it to be quite effective, several surrogate functions are identified in Section \ref{sec:surrogate}, and can instead be used, within our
detection framework.
Fourth, $\delta$ is the step size for updating $\underline{v}$. If $\delta$ is too small, the execution time can be very long. If $\delta$ is too large, the algorithm may terminate in a few steps with a resulting $\rho$ much larger than $\pi$ and with the resulting perturbation much larger in size/energy than that required to induce $\pi$-level group misclassification. In practice, a suitable $\delta$ can be chosen via line search. Fifth, in our actual implementation, we set an upper bound on the size/energy of $\underline{v}$, with the algorithm terminated when the upper bound is reached even if the $\pi$ target has not been reached -- when $\pi$ is selected too large (e.g. $\pi=1$), looking for the required perturbation may be hard or even infeasible (even for a ground-truth attacked pair $(s^{\ast}, t^{\ast})$).

\subsubsection{Detection Inference}\label{sec:inference}

The statistics for the detection inference are $\{r_{st}\}$, where $r_{st}=d(\underline{v}_{st}^{\ast})^{-1}$, i.e. the reciprocal of the ``size'' of the $K(K-1)$ optimized perturbations (one for each class pair). $d(\cdot)$ can {\it e.g.} be the L1 norm or L2 norm. We take the reciprocal because otherwise anomalies (corresponding to $(s^{\ast}, t^{\ast})$ or $(s, t^{\ast})$ for $\forall s\in\mathcal{S}^{\ast}$), if they exist, will be near the origin. The null hypothesis for our detection inference is that the classifier has not been attacked, which requires the $K(K-1)$ detection statistics to follow a null distribution (i.e. a distribution for non-backdoor attack pairs). Alternatively, if the classifier has been attacked, the reciprocal statistics corresponding to the class pairs involved in the attack should be large anomalies, with small p-values under the null. Hence our detection inference for each classifier consists of the following two steps: i) estimating a null distribution for non-backdoor class pairs; ii) evaluating whether the largest detection statistic, which is the one most likely associated with a backdoor attack, is very unlikely under the null distribution.

The null parametric density form used in our detection experiments is a Gamma distribution, a right-tailed distribution with positive support. We note that other one-tailed distributions, e.g. exponential distribution, inverse Gaussian distribution, etc., can also be used as the null distribution within our approach. Because we assume that there is at most one target class, if there is a backdoor attack, there can be at most $K-1$ reciprocal statistics that correspond to backdoor attack class pairs. Thus, we consider the $K(K-1) - (K-1) = (K-1)^2$ {\it smallest} reciprocals (corresponding to the $(K-1)^2$ optimized perturbations with the largest size/energy) and assume that these are {\it not} associated with backdoor attacks and thus are suitable for use in learning the null density function. We exclude the $(K-1)$ {\it largest} reciprocals from being used for such estimation because these could be associated with the backdoor attack and thus could corrupt estimation of the null model, as will be shown experimentally in Section \ref{sec:OtherExp}. However, one should not simply {\it naively} learn a null density using the $(K-1)^2$ smallest reciprocal
statistics. Note that it is {\it unknown} which, if any, of the $(K-1)$ largest reciprocals correspond to a backdoor -- there may be no backdoor attack on the network. Thus, it is incorrect to assume there are {\it no} null measurements in the interval $r_{\rm min} < r < r_{\rm max}$, where $r_{\rm min}$ is the smallest of the $(K-1)$ largest reciprocals and $r_{\rm max}$ is the largest of these statistics -- some of these $(K-1)$ statistics could correspond to class pairs not involved in a backdoor attack and follow the true null distribution. Even if we do not use these observed statistics to estimate the null, we should not implicitly assume there are no null measurements in this interval. To account for this lack of certainty, we should use the $(K-1)^2$ smallest statistics to learn the {\it conditional} null density, where we condition on these statistics being less than $r_{\rm min}$. That is, we condition on the observed statistics being smaller than the {\it smallest} of the $(K-1)$ statistics that could correspond to a backdoor attack. Thus, we use the $(K-1)^2$ smallest statistics to learn the conditional null density $g_R(r | r < r_{\rm min})$ by maximum likelihood estimation (MLE). Once we estimate this conditional density, its parameters then uniquely {\it determine} the corresponding {\it unconditional} null density $g_R(r)$. That is, $g_R(r)=g_R(r | r < r_{\rm min})\cdot {\rm prob}[R < r_{\rm min}]$, $r\geq 0$. To our knowledge, this is a novel robust density modeling approach we are proposing. In Section \ref{sec:all_stats}, we will demonstrate the superiority of this learned null model, compared with the naive null obtained by directly estimating the (unconditioned) null distribution using all the $K(K-1)$ reciprocal statistics.

\begin{figure}[h]
	\vspace{-0.15in}
	\begin{minipage}{1.0\linewidth}
		\centering
		\centerline{\includegraphics[width=6.7cm]{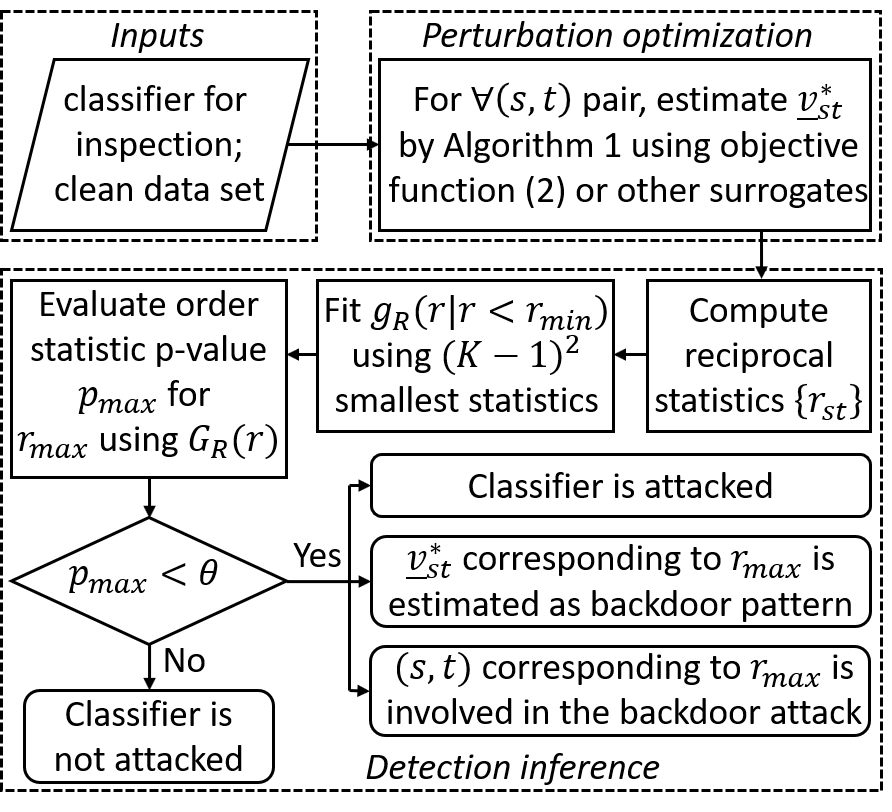}}
	\end{minipage}
	\caption{Flow chart for our complete detection procedure.}
	\label{fig:flowchart}
	\vspace{-0.25in}
\end{figure}

Given this learned null, we then ascertain if any of the $K(K-1)$ reciprocals deviate from the null. In particular, we evaluate the probability under the null that the largest of these $K(K-1)$ reciprocals is greater than or equal to the observed maximum reciprocal, $r_{\rm max}$, i.e.
\begin{equation*} \tag{4}\label{eq:order_cdf}
\begin{aligned}
{\rm p}_{\rm max} &\coloneqq {\rm prob}_{\rm null}[{\rm max}\{R_1, \ldots, R_{K(K-1)}\}\geq r_{\rm max}]\\
& =1-{\rm prob}_{\rm null}[{\rm max}\{R_1, \ldots, R_{K(K-1)}\}\leq r_{\rm max}]\\
& =1-G_R(r_{\rm max})^{K(K-1)}
\end{aligned}
\end{equation*}
where $G_R(\cdot)$ is the null cumulative distribution function. If this order statistic p-value is less than a threshold $\theta$, the null hypothesis is rejected and the classifier is claimed to be attacked. Since p-values under the null hypothesis are uniformly distributed on $[0, 1]$, $\theta$ can in principle be set to fix the false detection rate. For example, the ``classical'' statistical significance threshold $\theta=0.05$ should induce $5\%$ false detections for classifiers not attacked. If an attack is detected, the class pair corresponding to $r_{\rm max}$ is inferred to be involved in the backdoor attack. Furthermore, if the attack is detected, the optimized perturbation for the detected pair is our estimation of the backdoor pattern $\underline{v}^{\ast}$ used by the attacker. The {\it complete} detection procedure (which fits into the ``Backdoor detection'' block in Figure \ref{fig:post_training}), including this robust null learning and inference strategy, is summarized pictorially in Figure \ref{fig:flowchart}. Note that in \cite{ICASSP20}, we proposed a different order statistic inference approach, which is also shown to be quite effective.

\vspace{-0.1in}
\subsection{Surrogate Objective Function Variants}\label{sec:surrogate}

As noted above, $J_{st}(\cdot)$ measures a non-zero gradient contribution even from samples already successfully misclassified.  This can be remedied by instead minimizing an objective function similar 
to that associated with the Perceptron algorithm \cite{Duda}:
\begin{equation*} \tag{5}\label{eq:opt_perceptron}
\vspace{-0.05in}
J_{st-{\text{p}}}(\underline{v}) = -\frac{1}{|\hat{\mathcal{D}}_s(\underline{v}, t)|}\sum_{\underline{x}\in\hat{\mathcal{D}}_s(\underline{v}, t)} p(t|[\underline{x}+\underline{v}]_c),
\end{equation*}
where $\hat{\mathcal{D}}_s(\underline{v}, t) = \{\underline{x}\in\mathcal{D}_s: f([\underline{x}+\underline{v}]_c)\neq t\}$.

Another potential concern with $J_{st}(\underline{v})$ in Eq. (\ref{eq:obj_basic}) is the use of all images from source class $s$ for perturbation optimization -- if the attacker knows the training set and training approach, he/she can mimic (clean) classifier training and identify the training samples from $s$ that are misclassified.  It is possible the attacker will exclude these samples from consideration in crafting the backdoor. Accordingly, to mimic the attacker, the defender might consider group
misclassification only on the subset of clean source samples that are correctly classified.  This
leads to the following objective:
\begin{equation*} \tag{6}\label{eq:opt_correct}
J_{st-{\text{c}}}(\underline{v}) = -\frac{1}{|\hat{\mathcal{D}}_s|}\sum_{\underline{x}\in\mathcal{D}_s, f(\underline{x})=s} p(t|[\underline{x}+\underline{v}]_c),
\vspace{-0.075in}
\end{equation*}
where $\hat{\mathcal{D}}_s = \{\underline{x}\in\mathcal{D}_s: f(\underline{x})=s\}$.
One can also combine the restrictions in (\ref{eq:opt_perceptron}) and (\ref{eq:opt_correct}),
summing only over $\underline{x}$ such that both $f([\underline{x}+\underline{v}]_c)\neq t$ and $f(\underline{x})=s$.

Likewise, one might also hypothesize the attacker used box constraints \cite{Szegedy_seminal}, rather than clipping, in implementing the backdoor to keep image intensity values in the proper range.  If so,
the defender might modify the above surrogate problems for box constraints, rather than clipping.

As noted before, one could also consider an optimization problem that explicitly accounts for/constrains
the perturbation size.
Moreover, one can use a soft error count/discriminative learning objective function akin to that used
in \cite{Juang}.  

Based on this discussion, one can see that many surrogative objective function variants are possible.
In Section \ref{sec:experiments}, we experiment with some of these alternative surrogates; 
however, in general we have found that detection performance is not sensitive 
to this choice. Again, the reason is that all that is needed is that the size of the learned perturbation for a true backdoor pair $(s^{\ast}, t^{\ast})$ should be much smaller than for a non-backdoor pair. Minimizing any of these surrogate objectives is sufficient to elicit this difference between $r_{s^{\ast}t^{\ast}}$ and $r_{st}$, $(s, t)$ any non-backdoor pair. While our experiments primarily focus on minimizing $J_{st}(\cdot)$, our detection framework is general and consistent with use of any of the above alternative surrogates, in the search for minimal-sized image perturbations.

\vspace{-0.1in}
\subsection{Correction for Class Confusion}\label{sec:correctness}

The premise behind our detection inference is that the perturbation size/energy required to induce $\pi$-level group misclassification for $(s^{\ast}, t^{\ast})$ is less than for other
(unattacked) class pairs. However, if the initial misclassification fraction $\rho_{st}^{(0)}$ for $(s, t)\neq(s^{\ast}, t^{\ast})$ is abnormally high, it is expected that the perturbation size/energy needed to reach $\pi$-level group misclassification will be smaller than when this initial misclassification fraction is low, i.e. $d(\underline{v}_{st}^{\ast})$ may be abnormally small, possibly resulting in $(s, t)$ being falsely detected.
\begin{figure}[h]
	\vspace{-0.15in}
	\begin{minipage}{1.0\linewidth}
		\centering
		\centerline{\includegraphics[width=7.5cm]{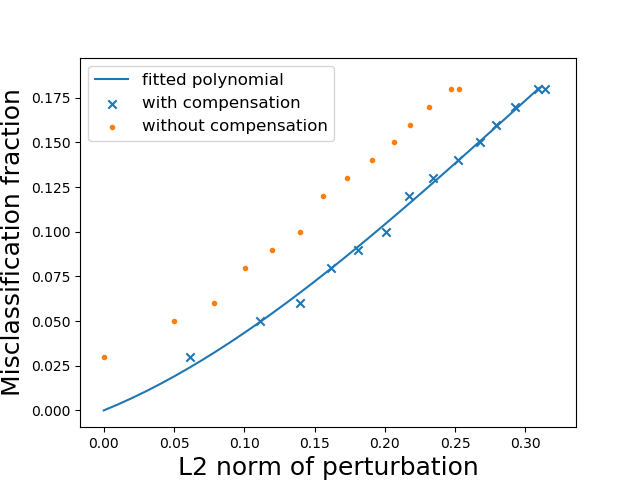}}
	\end{minipage}
	\caption{An example of correction of perturbation sizes to account for non-zero initial class pair confusion.}
	\label{fig:compensation}
	\vspace{-0.2in}
\end{figure}

Here we suggest to correct this effect assuming the confusion matrix information is available (estimable from the available clean data set). For any pair $(s, t)$ such that $\rho_{st}^{(0)}\neq 0$, we fit an (M-th order) polynomial using the sequence of (perturbation size/norm, misclassification fraction) pairs obtained while executing Algorithm \ref{alg:main} for the pair $(s,t)$. 
This gives a regression relationship between perturbation size and induced misclassification fraction for
the pair $(s,t)$.
The ``compensation'' $d_{st}^{(0)}$ is derived as
\begin{equation*} \tag{7}\label{eq:compensation}
d_{st}^{(0)} = \argmin_{d_0} \min_{\{a_m\}} \sum_{\tau=0}^{T} \big[\sum_{m=1}^M a_m (d(\underline{v}_{st}^{(\tau)})+d_0)^m -\rho_{st}^{(\tau)}\big]^2,
\end{equation*}
where the superscript $\tau$ is the index of the iteration when executing Algorithm \ref{alg:main}. 
$d_{st}^{(0)}$ is an estimate of the perturbation size needed to induce the initial class confusion
$\rho_{st}^{(0)}$.  To correct for this (non-zero) initial confusion level, we thus add $d_{st}^{(0)}$
to $d(\underline{v}_{st}^{\ast})$.  Intuitively, this correction will increase an ``abnormally small'' $d(\underline{v}_{st}^{\ast})$, making it less likely to be falsely detected.
Note that $d(\underline{v}_{st}^{(0)}) = 0$ since $\underline{v}_{st}^{(0)} = \underline{0}$.\footnote{In Algorithm \ref{alg:main}, we neglect the subscripts for simplicity.} $T$ could be chosen as the index of the last iteration such that either $\rho_{st}^{(T)}\geq\pi$ or such that $d(\underline{v}_{st}^{(T)})$ exceeds the upper bound of the perturbation size/norm. 
Alternatively, we can base the polynomial fit on
a smaller number of (perturbation size, misclassification rate) pairs.  For $(s, t)$ with $\rho_{st}^{(0)}= 0$, we set $d_{st}^{(0)}=0$, i.e. no correction is required in this case.

Figure \ref{fig:compensation} gives an example of correction to $d(\underline{v}_{st}^{\ast})$ for one class pair using the confusion information. The orange dots are the original (perturbation size/norm, misclassification fraction) sample points obtained from Algorithm \ref{alg:main}. The compensation based on an optimized polynomial with order $M=3$ (the blue curve in the figure) is obtained by solving (\ref{eq:compensation}). The blue crosses are the sample points after correction. Note that the polynomial is guaranteed to pass through the origin, since the purpose of the correction is to ensure that the initial misclassification rate when there is no perturbation is zero. Then the corrected statistics for detection inference are $\{r_{st}^c\}$, where $r_{st}^c = [d(\underline{v}_{st}^{\ast})+d_{st}^{(0)}]^{-1}$.

\vspace{-0.1in}
\section{Related Work}\label{sec:related_work}

Few existing works seek to thwart backdoor attacks post-training. Fine-pruning (FP) was proposed in \cite{FP}, where the defender prunes neurons in the penultimate layer in increasing order of their average activations over the clean data set, doing so up until the point where there is an unacceptable loss in classification accuracy on the clean data set. The premise behind FP is that backdoor patterns will activate neurons that are not triggered by clean examples and pruning will likely remove these neurons. However, this premise is not so plausible in general because there is nothing inherent to gradient-based neural net training (on a poisoned training set) that would create a propensity for a simple ``dichotomization'' of neurons, with most {\it solely} dedicated to ``normal'' operation and some {\it solely} dedicated to implementing the backdoor.

Neural Cleanse (NC) was one of the first post-training detection methods that involves perturbation optimization and anomaly detection, like our approach \cite{NC}. NC assumes that the backdoor pattern $\underline{v}^{\ast}$ is embedded into a clean image $\underline{x}$ by
\begin{equation*}\tag{8}\label{eq:bd_embedding_perceptible}
B_{\rm mask}(\underline{x}, \underline{v}^{\ast}, \underline{m}^{\ast}) = \underline{x}\odot(\underline{1}-\underline{m}^{\ast})+\underline{v}^{\ast}\odot\underline{m}^{\ast},
\vspace{-0.05in}
\end{equation*}
where $\underline{m}^{\ast}$ is a 2-dimentional, image-wide mask, with values between 0 and 1; $\odot$ is element-wise multiplication that applies the mask $\underline{m}^{\ast}$ (or its ``inverse'' $(\underline{1}-\underline{m}^{\ast})$) to each channel/color of the image. NC further assumes that: 1) the backdoor pattern should have a small spatial support, i.e. the mask $\underline{m}^{\ast}$ should have as few non-zero elements as possible; 2) if there is an attack, {\it all} classes except for the target class are source classes. Hence, for each putative target class $t$, NC jointly estimates a pattern and a mask using clean images from all classes except for class $t$ by minimizing the following Lagrangian:
\begin{equation*}\tag{9}\label{eq:obt_NC}
\vspace{-0.05in}
L_{\rm NC} (\underline{v}, \underline{m}) = - \frac{1}{|{\mathcal D} \setminus {\mathcal D}_t|} \sum_{\underline{x}\in {\mathcal D} \setminus {\mathcal D}_t} p(t|B_{\rm mask}(\underline{x}, \underline{v}, \underline{m})) + \lambda \cdot ||\underline{m}||_1,
\end{equation*}
with $\lambda$ prescribed, until a high misclassification rate to class $t$ from all the other classes is achieved. Then anomaly detection is performed on the L1 norm of the $K$ estimated masks, each for a putative target class. An ``anomaly index'' is derived for each class using median absolute deviation (MAD) \cite{MAD}. If for any class $t$, the anomaly index is greater than a threshold $\theta_{\rm MAD}=2$, a detection is made with $>95\%$ confidence, with $t$ the target class \cite{NC}.

Unfortunately, the assumptions made by NC largely limits its usability/performance in practice. First, an attacker may avoid choosing too many source classes when devising the attack; otherwise, too many examples with the backdoor pattern will be inserted in the training set, making the attack less evasive. In Section \ref{sec:eval_main}, we will show that NC fails when there is a single source class involved in the attack. Second, NC generates $K$ (compared with $K(K-1)$ for our approach) statistics for detection inference, such that (assuming a single target class of the attack) there are only $(K-1)$ ``null value realizations'' (decision statistics guaranteed to {\it not} be associated with an attack), against which to assess atypicality of the most extreme statistic. This may lead to an unreliable inference result. Third, as shown by both our experiments and in \cite{Tabor}, when there is no attack, NC can still estimate a pattern with an abnormally small spatial support for some classes, resulting in a false detection. Fourth, NC was only experimentally evaluated against perceptible attacks -- the noisy backdoor pattern used in their experiments is easy visibly discernible, unlike the imperceptible backdoor patterns considered in this paper.

A recent extension of NC, named ``TABOR'', was proposed in \cite{Tabor}. It focuses on detecting a very special subset of backdoor patterns located in any one of the four corners of the image. TABOR assumes that the spatial support of the pattern should be very small and non-scattered. It also assumes that the backdoor pattern should not cover any key features in the image. The (pattern, mask) estimation of TABOR uses the same objective function as NC, added with regularization terms penalizing any (pattern, mask) that violates its assumptions. Compared with TABOR (and NC), our AD approach does not constrain the shape, location, or spatial support of the backdoor pattern. In Section \ref{sec:experiments}, we will show that sparse pixel-wise perturbations and global, image-wide perturbations with small perturbation size are both effective as backdoor patterns -- our AD detects these attacks with high accuracy.

\vspace{-0.1in}
\section{Experiments}\label{sec:experiments}
\vspace{-0.1in}
\subsection{Backdoor Patterns}
We first focus on two types of backdoor patterns based on additive image perturbation. Since pixel values are usually normalized (e.g. from $[0, 255]$ for colored images) to $[0, 1]$ before feeding to DNN classifiers, the valid perturbation range per pixel (without clipping) is $[-1, 1]$. The first type is the sparse pixel-wise perturbation, a backdoor pattern that affects only a small subset of pixels, as considered in \cite{Song}\cite{Trojan}\cite{AC}\cite{BadNet}\cite{SS}\cite{NC}\cite{FP}\cite{Tabor}. Our pixel-wise perturbation is created by first randomly selecting a few pixels; for colored images, one of the three channels (i.e. colors) of each selected pixel is randomly selected to be perturbed. The perturbations can be either positive or negative, but the perturbation magnitude is similar, across all the chosen pixels. In our experiments, this was achieved by first fixing a ``reference'' perturbation magnitude and then multiplying this reference magnitude for each of the chosen pixels by a random factor generated by a Gaussian distribution with mean 1 and standard deviation 0.05. The second type of pattern is a global, i.e. image-wide, perturbation, a backdoor pattern that affects all pixels, akin to a global image watermark as considered in \cite{Song}\cite{Haoti}\cite{Hidden-trigger}. We created a spatially recurrent pattern that looks like a ``chess board'' -- one and only one pixel among any two adjacent pixels was perturbed (in all three channels) positively. Again, the perturbation magnitude for each pixel being perturbed is a fixed value multiplied by a random factor generated from the same Gaussian distribution mentioned above. Finally, if the L2 norm of the perturbation is specified, we can always scale the perturbation mask to meet the specification.

\begin{figure}[h]
	\vspace{-0.15in}
	\begin{minipage}[b]{.48\linewidth}
		\centering
		\centerline{\includegraphics[width=4cm]{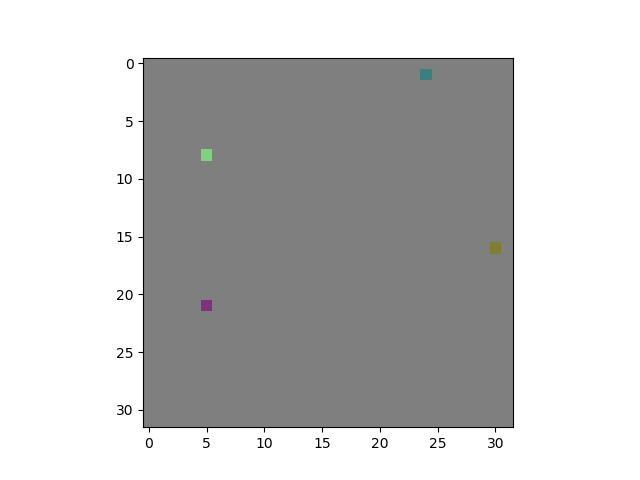}}
	\end{minipage}
	\hfill
	\begin{minipage}[b]{0.48\linewidth}
		\centering
		\centerline{\includegraphics[width=4cm]{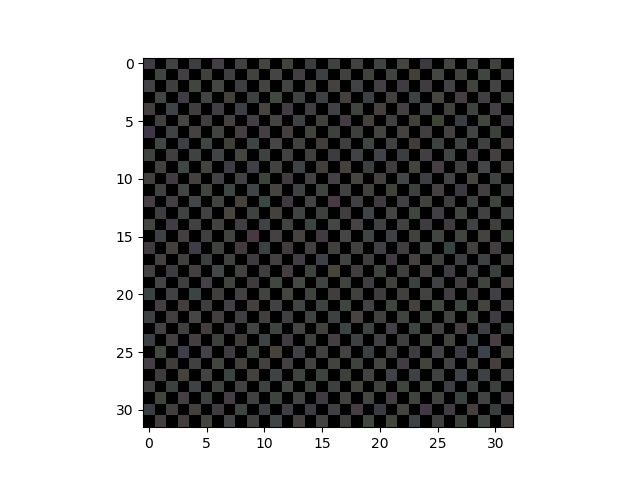}}
	\end{minipage}
	\caption{A sparse pixel-wise perturbation mask with L2 norm 0.6 and a 0.5 offset (left), and a global perturbation mask with L2 norm 10 (right).}
	\label{fig:perturbation}
	\vspace{-0.15in}
\end{figure}
Examples of the two types of backdoor patterns for colored images are shown in Figure \ref{fig:perturbation}. On the left, we show the backdoor pattern from one attack realization used in the experiments in Section \ref{sec:ensemble}. Four pixels were randomly selected to be perturbed (in one of the three channels), with the L2 norm\footnote{There is no strong preference on the norm to be used. L2 norm is the default, unless specified otherwise. However, we will also evaluate our detection algorithm using the L1 norm.} of the perturbation mask equal to 0.6. For visualization purpose, a 0.5 offset was added to each pixel of the image. On the right, a global perturbation mask with L2 norm equal to 10 is shown. However, to launch a successful backdoor attack with a global backdoor pattern, a perturbation mask with such large L2 norm is unnecessary -- the required norm is in fact smaller than for sparse attacks.

\vspace{-0.1in}
\subsection{Ensemble Experiments}\label{sec:ensemble}

In this experiment, we evaluate performance of the proposed defense and some of its variants (i.e. surrogate objective functions and confusion-corrected detection inference described in Section \ref{sec:detection}) based on multiple realizations of classifiers under different attack settings and the two types of backdoor patterns mentioned above. When doing experiments, we of course have omniscient knowledge of the classifier and the data poisoning, etc. Thus, as advocated in \cite{Review}, we are very careful, in doing experiments, to compartmentalize this knowledge when implementing the defender and the attacker, so that the defense and attack only exploit information consistent with their stated assumptions. Our experiment uses the CIFAR-10 data set with 60k color images ($32\times 32\times 3$) evenly distributed between ten classes. The data set is separated into a training set with 50k images (5k per class) and a test set with 10k images (1k per class). The DNN classifier uses the ResNet-20 \cite{ResNet} structure with cross-entropy training loss. Training is performed for 200 epochs with mini-batch size of 32, using the Adam optimizer \cite{Adam}, and with training data augmentation (see supplementary material for details). This training achieves an accuracy of 91\% on the clean test set when there are no backdoor attacks. Note that this data set, DNN structure and training configurations are frequently used in many adversarial learning works. Experiments on other data sets and DNN structures can be found in the supplementary material.

\subsubsection{Devising Backdoor Attacks}

\begin{figure}[h]
	\begin{minipage}{1.0\linewidth}
		\vspace{-0.1in}
		\centering
		\centerline{\includegraphics[width=9cm]{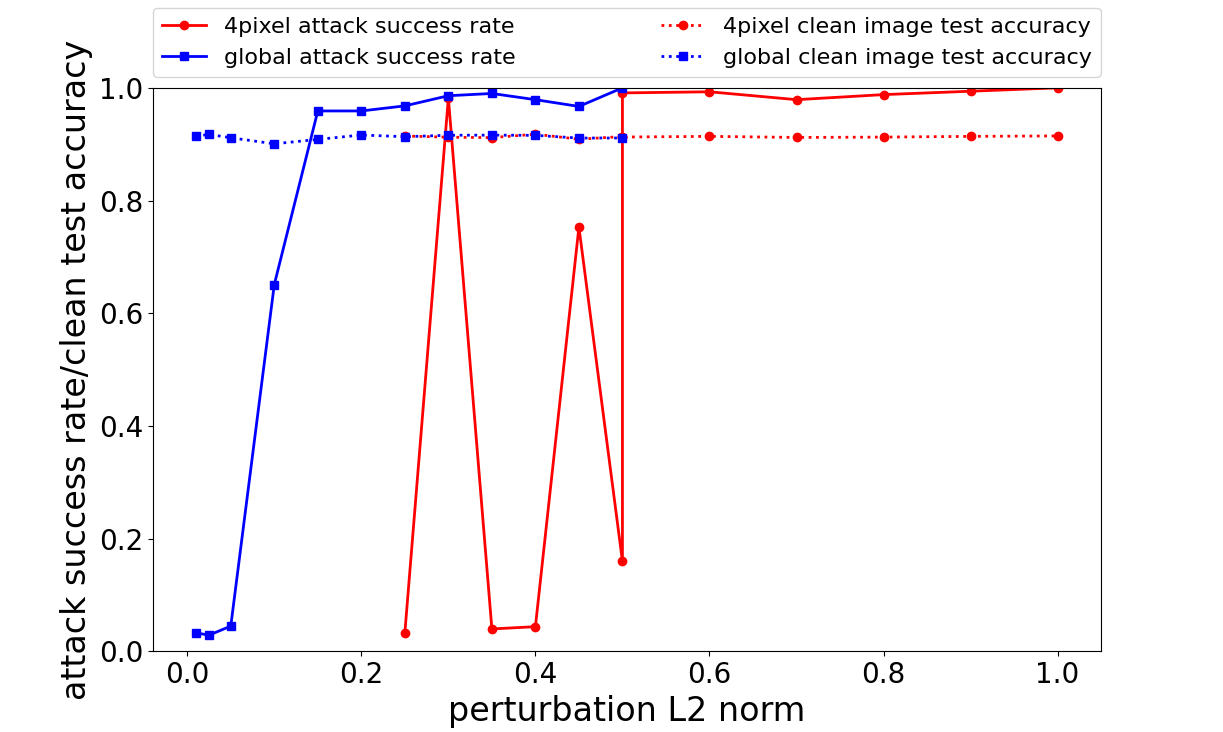}}
	\end{minipage}
	\caption{Attack success rate (solid) and accuracy on clean test images (dashed) for DNN classifiers under backdoor attacks that use sparse pixel-wise perturbation (red dots) and global perturbation (blue squares), for a range of L2 perturbation norms (attack strengths).}
	\label{fig:attack_succ}
	\vspace{-0.125in}
\end{figure}

Consistent with our description in Section \ref{subsec:attacker_assump}, we evaluated perturbations (i.e. the backdoor pattern) with different L2 norms, and picked the least human-perceptible one (as small L2 norm as possible) under the constraints of high attack success rate and low degradation in test accuracy on clean patterns. Such evaluation is given in Figure \ref{fig:attack_succ}. A larger perturbation norm could be used but at the risk of being perceivable to humans. Also, our detector can still detect such attacks, as will be seen in Section \ref{sec:size_of_gt_bd}. For sparse pixel-wise perturbations and global perturbations, we evaluated L2 perturbation norms ranging from 0.25 to 1.0 and 0.01 to 0.5, respectively. For each L2-norm-specified backdoor pattern, we created a single attack realization. We produced 1000 attack images using randomly selected\footnote{The attacker could alternatively select the images that are easiest to be misclassified to the target class, based {\it e.g.} on the difference in DNN posterior probability between the winning class and the target class.} clean training images from the `automobile' (source) class, adding the backdoor pattern to each image, and then clipping as described by Eq. (\ref{eq:bd_embedding_imperceptible}). These images (with the backdoor pattern) were labeled to the `truck' (target) class and added to the training set of 50k clean images. The poisoned training set was then used to train a DNN classifier. A set of backdoor test patterns was created by adding the same backdoor pattern to the 1k clean test patterns (those not used in training) from the source class. The attack success rate is evaluated as the fraction of backdoor test patterns classified to the target class. Also, the accuracy of the classifier on the original 10k clean test patterns is evaluated.

As depicted in Figure \ref{fig:attack_succ}, for the global perturbation backdoor pattern, the attack success rate grows with the L2 norm of the perturbation mask. However, for the sparse, pixel-wise backdoor pattern, the attack success rate wildly fluctuates when the perturbation norm is low, only becoming stable with further increases in the perturbation norm. Such fluctuation is likely due to the fact that the pixels being perturbed are randomly selected. Consider an attack realization in which the neighborhood of a perturbed pixel is noisy for most of the selected images. Such a backdoor pattern might be poorly learned by the classifier. Based on the results shown in Figure \ref{fig:attack_succ}, to launch a successful and human-imperceptible attack, the attacker should choose the L2 norm for the global perturbation mask to be 0.2, with an attack success rate of 0.959 and test set accuracy of 0.916 on clean test images. For the sparse, pixel-wise perturbation mask, if the L2 norm is set to 0.6, an attack success rate of 0.993 and test accuracy of 0.914 on clean test images can be achieved.

\begin{figure}[h]
	\vspace{-0.1in}
	\centering
	\begin{minipage}[b]{.3\linewidth}
		\centering
		\centerline{\includegraphics[width=3.4cm]{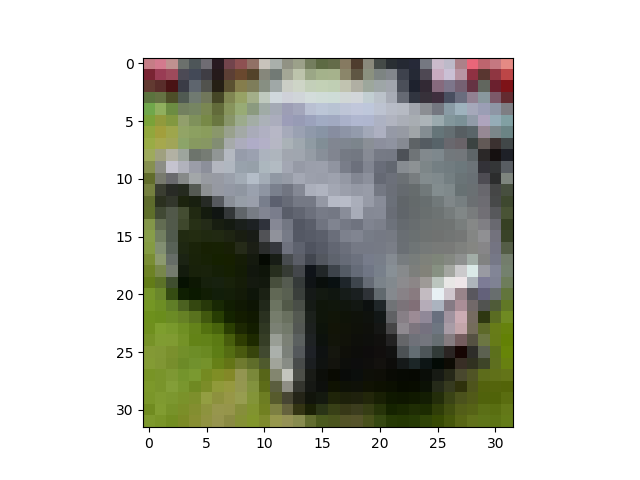}}
		\subcaption{}\label{fig:auto_before}
	\end{minipage}
	\begin{minipage}[b]{0.3\linewidth}
		\centering
		\centerline{\includegraphics[width=3.4cm]{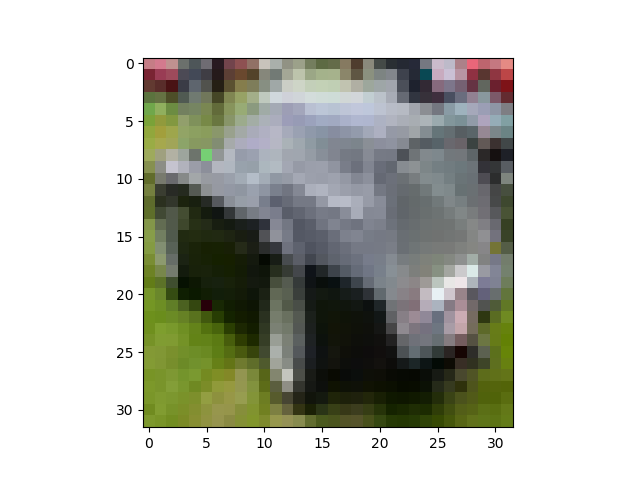}}
		\subcaption{}\label{fig:auto_4pixel}
	\end{minipage}
	\begin{minipage}[b]{.3\linewidth}
		\centering
		\centerline{\includegraphics[width=3.4cm]{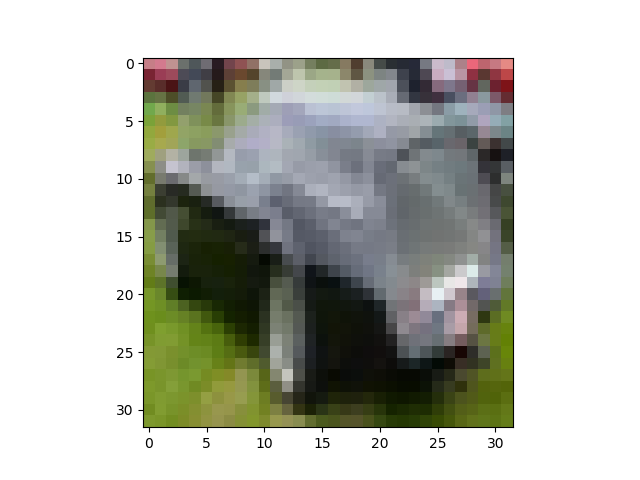}}
		\subcaption{}\label{fig:auto_glb}
	\end{minipage}
	\caption{Examples of backdoor patterns applied to CIFAR-10 images: (a) the original automobile image; (b) automobile with sparse, pixel-wise perturbation ($||\underline{v}^{\ast}||_2=0.6$); (c) automobile with global perturbation ($||\underline{v}^{\ast}||_2=0.2$).}
	\label{fig:before_after}
	\vspace{-0.15in}
\end{figure}

Based on the randomness in crafting the backdoor patterns and in the training process\footnote{In each training iteration, the mini-batch is randomly sampled from the training set. Also, for each training image, the type of data augmentation is randomly chosen.}, to evaluate detection performance, we conducted experiments on four groups of DNN classifiers, 25 classifier realizations per group, as follows:
\begin{itemize}
	\item {\bf BD-P-S}: Classifiers were trained on the training set poisoned by 1000 backdoor images with 4-pixel perturbations ($||\underline{v}^{\ast}||_2=0.6$). The backdoor images were crafted using clean images from a single source class.
	\item {\bf BD-G-S}: Classifiers were trained on the training set poisoned by 1000 backdoor images with a global perturbation ($||\underline{v}^{\ast}||_2=0.2$). The backdoor images were crafted using clean images from a single source class.
	\item {\bf BD-G-M}: Classifiers were trained on the training set poisoned by 900 backdoor images with a global perturbation ($||\underline{v}^{\ast}||_2=0.2$). The backdoor images were crafted using clean images from the nine classes excluding the target class, with 100 images per (source) class.
	\item {\bf Clean}: Classifiers were trained on the clean training set, without data poisoning.
\end{itemize}
The four groups of classifiers involve sparse, pixel-wise and global backdoor patterns, single-source-class and multiple-source-class attack scenarios, and include a clean classifier group. For each attacked classifier, the clean images used for devising the attack are selected randomly from the `automobile' (source) class, and the backdoor pattern is generated independently of the selected images. The attack success rate\footnote{The attack success rate for the BD-G-M group is evaluated on backdoor images crafted using all 9000 clean test images from the nine classes other than the target class.} and the clean test accuracy (across the 25 realizations) for the four groups of classifiers are reported in Table \ref{tab:4group} -- all backdoor attacks are successful and the degradation to clean test accuracy is negligible across all experimental realizations. An example of backdoor images with sparse, pixel-wise backdoor pattern (Figure \ref{fig:auto_4pixel}), global backdoor pattern (Figure \ref{fig:auto_glb}) and the original image (Figure \ref{fig:auto_before}) are shown in Figure \ref{fig:before_after}. The sparse, pixel-wise backdoor pattern (with L2 norm 0.6) is only human-perceived through careful visual scrutiny of the image. The global backdoor pattern (with L2 norm 0.2), when added to the original image, is imperceptible even under careful human inspection, since the perturbation size per pixel (and per channel) is only about $5.1\times 10^{-3}$.

\begin{table}[h]
	\vspace{-0.1in}
	\begin{center}
		\caption{Attack success rate and accuracy on clean test images (over all 25 classifier realizations) for the four groups of DNN classifiers.}
		\resizebox{0.40\textwidth}{!}{
			\begin{tabular}{ |c|c|c| } 
				\hline
				& Attack success rate & Test accuracy\\
				\hline
				BD-P-S & $0.978\pm 0.035$ & $0.913\pm 0.003$\\
				\hline
				BD-G-S & $0.974\pm 0.014$ & $0.912\pm 0.003$\\
				\hline
				BD-G-M & $0.990\pm 0.005$ & $0.912\pm 0.003$\\
				\hline
				Clean & N.A. & $0.915\pm 0.003$\\
				\hline
			\end{tabular}\label{tab:4group}}
	\end{center}
	\vspace{-0.15in}
\end{table}

\begin{figure}[h]
	\vspace{-0.05in}
	\centering
	\begin{minipage}[b]{.48\linewidth}
		\centering
		\centerline{\includegraphics[width=4.7cm]{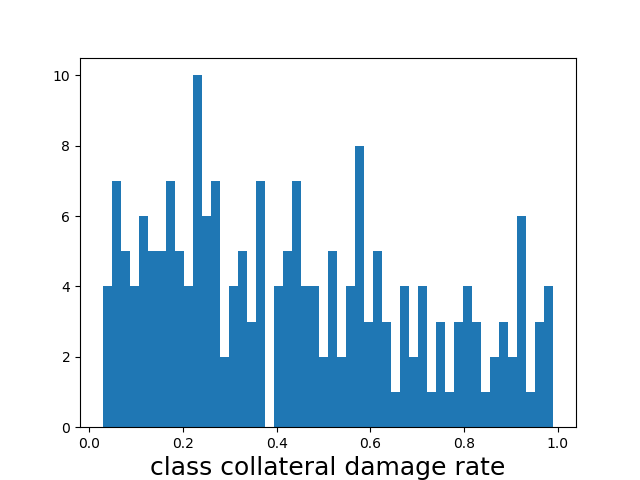}}
		\subcaption{BD-P-S group}\label{fig:collateral_4pixel}
	\end{minipage}
	\begin{minipage}[b]{0.48\linewidth}
		\centering
		\centerline{\includegraphics[width=4.7cm]{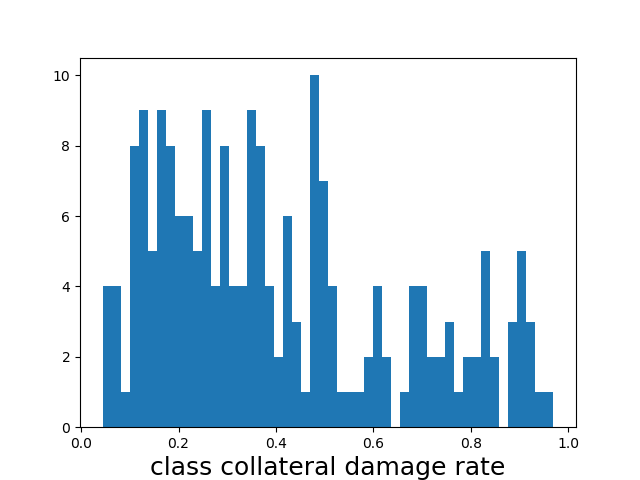}}
		\subcaption{BD-G-S group}\label{fig:collateral_glb}
	\end{minipage}
	\caption{Histograms of the collateral damage rate statistics for (a) the BD-P-S group and (b) the BD-G-S group.}
	\label{fig:collateral}
	\vspace{-0.3in}
\end{figure}

During the experiments, we noticed that backdoor DP attacks usually induce ``collateral damage'' to classes other than $s^{\ast}$ or $t^{\ast}$. That is, supposing the classifier has been successfully corrupted by a backdoor involving the pair $(s^{\ast}, t^{\ast})$, test images from some classes $\tilde{s}\in\mathcal{C}\setminus\{s^{\ast}, t^{\ast}\}$ will be classified to $t^{\ast}$ with high probability when the same backdoor pattern is added to them. To demonstrate this effect, for each trained DNN in the BD-P-S and BD-G-S groups, for which only one class is designated to be the source class $s^{\ast}$ of the backdoor, we conducted eight tests, one for each of the classes ($\tilde{s}$) other than $s^{\ast}$ or $t^{\ast}$. In each test, we added the backdoor pattern used for attack to the 1000 clean test images from $\tilde{s}$, and obtained a ``class collateral damage rate'' as the fraction of images classified to $t^{\ast}$. In Figure \ref{fig:collateral_4pixel} and Figure \ref{fig:collateral_glb}, we show the histogram of the ($8\times 25=$) 200 class collateral damage rate statistics for the BD-P-S and BD-G-S groups, respectively. Clearly, significant group misclassification to the target class is induced for classes other than the true source class using the true backdoor pattern, for many of the classifiers. 
This is explained by some classes ($\tilde{s}$) having patterns that are similar to those of the backdoor's source class -- the backdoor attack is not sufficiently ``surgical'' to {\it only} induce misclassifications from $s^{\ast}$ to $t^{\ast}$.
Accordingly, it is not  
surprising that, for such high ``collateral damage'' classes, the size/energy of the optimized $\underline{v}_{\tilde{s}t^{\ast}}^{\ast}$ is close to the size/energy of $\underline{v}_{s^{\ast}t^{\ast}}^{\ast}$. 
This thus gives further insight into why we exclude the $K-1$ largest statistics in the set $\{||\underline{v}_{st}||\}$ from use in estimating the null distribution.
This also helps explain why we only
infer a single (source, target) class pair based on the {\it most extreme} statistic (as illustrated in Figure \ref{fig:flowchart}) -- ``collateral damage'' classes, paired with the inferred target class, may be prone to false detection, and it is not so easy to distinguish the ``collateral damage'' phenomenon for a single source attack from a truly multiple-source backdoor attack. 
We do not view this as a true limitation of our detection method -- our method should be able to accurately identify multiple true source classes (using the second to the $(K-1)$-th largest order statistics) {\it if} the backdoor attack is more surgical, {\it i.e.} if it does not induce collateral damage to classes not explicitly attacked. 

\subsubsection{Detection Performance Evaluation}\label{sec:eval_main}

We first evaluate the performance of the proposed AD framework and compare with that of the NC approach, using the four groups of classifiers. The clean data set used for detection (for both AD and NC) for each classifier contains 1000 images (100 per class) randomly sampled from the clean test set held out from training. The combinations of the objective functions for perturbation optimization and the detection inference strategies to be evaluated are as follows:
\begin{itemize}
	\item {\bf AD-J-P}: Eq. (\ref{eq:obj_basic}) objective combined with the principal detection inference approach.
	\item {\bf AD-Jp-P}: Eq. (\ref{eq:opt_perceptron}) objective combined with the principal detection inference approach.
	\item {\bf AD-Jc-P}: Eq. (\ref{eq:opt_correct}) objective combined with the principal detection inference approach.
	\item {\bf AD-J-C}: Eq. (\ref{eq:obj_basic}) objective combined with the confusion-corrected detection inference approach described in Section \ref{sec:correctness}.
	\item {\bf AD-J-P-L1}: As the other variants take $d(\cdot)$ to be the L2 norm, this variant is the same as AD-J-P except using the L1 norm for detection purposes.
\end{itemize}
For all the variants above, we set $\pi$, the target misclassification fraction, to be 0.8. However, we have observed that detection performance is not sensitive to the choice of $\pi$ (shown in Section \ref{sec:choice_of_pi}). For the corrected detection inference used with the AD-J-C variant, we selected the order of the polynomial as $M=3$. Again, other choices of $M$ have little impact on detection accuracy -- $M=2, 4, 5$ gave similar results. The only remaining hyperparameter to be selected for all the variants is the detection threshold $\theta$ on the p-values. $\theta$ can be set to fix the theoretical false detection rate. Here we use the ``classical'' statistical significance threshold $\theta=0.05$ as the default, which also matches with the choice of NC in terms of the significance level. We will also evaluate performance for more conservative and liberal choices of $\theta$ later in the current section.

For NC, minimizing (\ref{eq:obt_NC}) requires specifying the Lagrange multiplier $\lambda$. We tested both $\lambda=1.5$ and $\lambda=1.0$.\footnote{The default is $\lambda=1.5$, see ``https://github.com/bolunwang/backdoor''.} We conjecture (without testing) that if $\lambda$ is selected too large, perturbation optimization for NC may never reach the $\pi$-level group misclassification target. For NC, we used the same $\pi=0.8$ target as for the proposed AD framework; we also evaluated $\pi=0.5$. However, we found for some classifiers that inducing $\pi$-level misclassification to a subset of putative target classes from all source classes (i.e. NC's objective) is not feasible. For these classes, the optimization process is terminated when the size/norm of the perturbation mask reaches a pre-set upper bound value. When this occurs, the derived detection statistics for these problematic classes skew estimation of the median during NC inference. In the experiments, we discarded these abnormal perturbation statistics in order to make NC inference more robust. For our proposed AD framework, this upper bound on the perturbation size/energy was never reached in our experiments -- $\pi$-level group misclassification was always achieved by our method. Even if the upper bound were to be reached for several class pairs, taking the reciprocal will send these large statistics close to zero and have little effect on the estimation of the tail of the null density. The variants of NC detection that we evaluated are:
\begin{itemize}
	\item {\bf NC-L1-1.5}: L1-regularized objective function with $\lambda=1.5$. The anomaly indices are derived from the L1 norm of each optimized perturbation.
	\item {\bf NC-L2-1.5}: L2-regularized objective function with $\lambda=1.5$. The anomaly indices are derived from the L2 norm of each optimized perturbation.
	\item {\bf NC-L1-1.0}: L1-regularized objective function with $\lambda=1.0$.
	\item {\bf NC-L1-1.5-0.5}: Same as NC-L1-1.5, except that $\pi=0.5$ instead of $\pi=0.8$.
\end{itemize}

\begin{table}[h]
	\begin{center}
		\caption{Detection accuracy of all the variants of the proposed AD framework (with detection threshold $\theta=0.05$) and of the NC approach for the four groups of DNN classifiers for CIFAR-10.}
		\resizebox{0.45\textwidth}{!}{
			\begin{tabular}{ |c|c|c|c|c| } 
				\hline
				& BD-P-S & BD-G-S & BD-G-M & Clean \\ 
				\hline
				AD-J-P & 0.92 & 0.92 & {\bf 1.00} & {\bf 1.00}\\
				\hline 
				AD-Jp-P & 0.92 & 0.92 & {\bf 1.00} & {\bf 1.00}\\
				\hline
				AD-Jc-P & 0.96 & 0.92 & {\bf 1.00} & {\bf 1.00}\\
				\hline 
				AD-J-C & 0.96 & 0.92 & {\bf 1.00} & {\bf 1.00}\\
				\hline 
				AD-J-P-L1 & {\bf 1.00} & 0.92 & {\bf 1.00} & {\bf 1.00}\\
				\hline
				\hline
				NC-L1-1.5 & 0.36 & 0.16 & {\bf 1.00} & 0.84\\
				\hline
				NC-L2-1.5 & 0.56 & 0.64 & {\bf 1.00} & 0.72\\
				\hline
				NC-L1-1.0 & 0.40 & 0.28 & {\bf 1.00} & 0.76\\
				\hline
				NC-L1-1.5-0.5 & 0.36 & 0.16 & 0.88 & 0.96\\
				\hline
			\end{tabular}\label{tab:result_main}}
	\end{center}
\vspace{-0.1in}
\end{table}

In Table \ref{tab:result_main}, detection accuracy of each variant of the proposed AD and NC detection methods for each group of classifiers is shown. Accuracy for the groups of classifiers under attack, i.e. BD-P-S, BD-G-S and BD-G-M, is defined as the fraction of classifiers successfully detected as attacked. For the proposed AD variants, a successful detection {\it also} requires the detected source and target class pair (corresponding to the most extreme detection statistic) to be ground-truth involved in the attack. For NC, since it assumes all classes except for the target class are involved in the backdoor attack, there is no inference on the source class(es). In our experiments, we thus consider NC detection to be successful if the class label corresponding to the most extreme anomaly index, if detected (i.e. above $\theta_{\rm MAD}$), is the true backdoor target class label $t^{\ast}$. 
Thus, for BD-P-S and BD-G-S, our AD approach meets a stronger true detection requirement than NC. For the group of clean networks, the detection accuracy is defined as the fraction of networks inferred to {\it not} be attacked.

All proposed AD variants achieved perfect detection for the BD-G-M and Clean group experiments. Detection accuracy for the BD-P-S and BD-G-S groups is also very high (all above 0.90). We note that we can always make more conservative AD inferences and claim a successful detection when {\it only} the target class is correctly inferred. This is reasonable due to the collateral damage effect shown previously. Allowing such conservative inference, all five AD variants achieve perfect detection for the BD-G-S group. In fact, all incorrect source classes by AD variants for the results shown in Table \ref{tab:result_main} have high class collateral damage rate (above $88.4\%$).

\begin{figure}[h]
	\vspace{-0.15in}	
	\begin{minipage}{1.0\linewidth}
		\centering
		\centerline{\includegraphics[width=7cm]{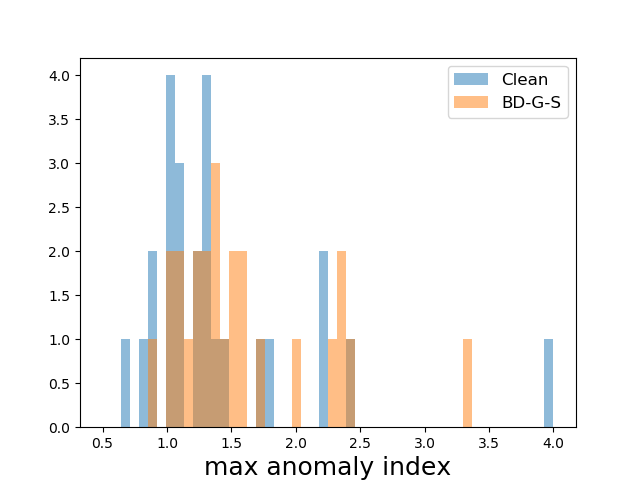}}
	\end{minipage}
	\caption{Histogram of the maximum anomaly indices for the BD-G-S and Clean groups, obtained using NC-L1-1.5.}
	\label{fig:NC_hist}
	\vspace{-0.15in}
\end{figure}

NC detection, as expected, achieves strong performance for the BD-G-M group (except for the NC-L1-1.5-0.5 variant) since it is designed for backdoor attacks that involve all $(K-1)$ source classes. However, all the variants of NC are not effective at detecting backdoor attacks involving a single source class, as shown in Table \ref{tab:result_main} by the accuracy for BD-P-S and BD-G-S. Also, the detection power of NC cannot be improved by choosing a smaller $\theta_{\rm MAD}$, since the false detection rate on clean classifiers is then made quite non-negligible. In Figure \ref{fig:NC_hist}, we show the histogram of the maximum anomaly indices for BD-G-S and Clean, obtained using NC-L1-1.5 -- these two groups of anomaly indices are not clearly separable by {\it any} choice of $\theta_{\rm MAD}$.

Unlike NC, for the proposed AD variants, the order statistic p-value for clean classifiers and for attacked classifiers are easily separable in practice -- there is a large range of thresholds $\theta$ which yield the same (or very similar) performance. In Table \ref{tab:result_main_0.2}, we show the detection performance for a very liberal detection threshold $\theta=0.2$. We can observe a slight increment in false detection rate for the Clean group. Simultaneously, the true detection rate for the BD-P-S group is slightly increased. If we use a very conservative detection threshold $\theta=0.01$, as shown in Table \ref{tab:result_main_0.01}, only the true detection rate for the BD-P-S group is reduced slightly. Moreover, we find from Table \ref{tab:result_main} and Table \ref{tab:result_main_0.2} that the false detection rate for the Clean group is much lower than its expectation, which equals the detection threshold $\theta$. This is because, for the classifiers in the Clean group, the distribution of the reciprocal statistics used for detection is not left-skewed like a typical Gamma distribution -- few reciprocals sit on the tail of the unconditional null distribution. Hence the order statistic p-value follows a slightly right-skewed distribution rather than a truly uniform distribution on $[0, 1]$. For example, the mean of the order statistic p-value across the 25 classifier realizations from the Clean group is 0.590 when the AD-J-P variant is applied for detection.

\begin{table}[h]
	\vspace{-0.05in}
	\begin{center}
		\caption{Detection accuracy of all the variants of the proposed AD framework with detection threshold $\theta=0.2$ for the four groups of DNN classifiers.}
		\resizebox{0.4\textwidth}{!}{
			\begin{tabular}{ |c|c|c|c|c| } 
				\hline
				& BD-P-S & BD-G-S & BD-G-M & Clean \\ 
				\hline
				AD-J-P & 0.96 & 0.92 & {\bf 1.00} & {\bf 1.00}\\
				\hline 
				AD-Jp-P & 0.96 & 0.92 & {\bf 1.00} & 0.96\\
				\hline
				AD-Jc-P & 0.96 & 0.92 & {\bf 1.00} & 0.96\\
				\hline 
				AD-J-C & 0.96 & 0.92 & {\bf 1.00} & {\bf 1.00}\\
				\hline 
				AD-J-P-L1 & {\bf 1.00} & 0.92 & {\bf 1.00} & 0.92\\
				\hline
			\end{tabular}\label{tab:result_main_0.2}}
	\end{center}
	\vspace{-0.15in}
\end{table}

\begin{table}[h]
	\vspace{-0.05in}
	\begin{center}
		\caption{Detection accuracy of all the variants of the proposed AD framework with detection threshold $\theta=0.01$ for the four groups of DNN classifiers.}
		\resizebox{0.4\textwidth}{!}{
			\begin{tabular}{ |c|c|c|c|c| } 
				\hline
				& BD-P-S & BD-G-S & BD-G-M & Clean \\ 
				\hline
				AD-J-P & 0.84 & 0.92 & {\bf 1.00} & {\bf 1.00}\\
				\hline 
				AD-Jp-P & 0.88 & 0.92 & {\bf 1.00} & {\bf 1.00}\\
				\hline
				AD-Jc-P & 0.88 & 0.92 & {\bf 1.00} & {\bf 1.00}\\
				\hline 
				AD-J-C & 0.96 & 0.92 & {\bf 1.00} & {\bf 1.00}\\
				\hline 
				AD-J-P-L1 & {\bf 1.00} & 0.92 & {\bf 1.00} & {\bf 1.00}\\
				\hline
			\end{tabular}\label{tab:result_main_0.01}}
	\end{center}
	\vspace{-0.25in}
\end{table}

\begin{figure}[h]
	\vspace{-0.15in}
	\centering
	\begin{minipage}[b]{.45\linewidth}
		\centering
		\centerline{\includegraphics[width=4cm]{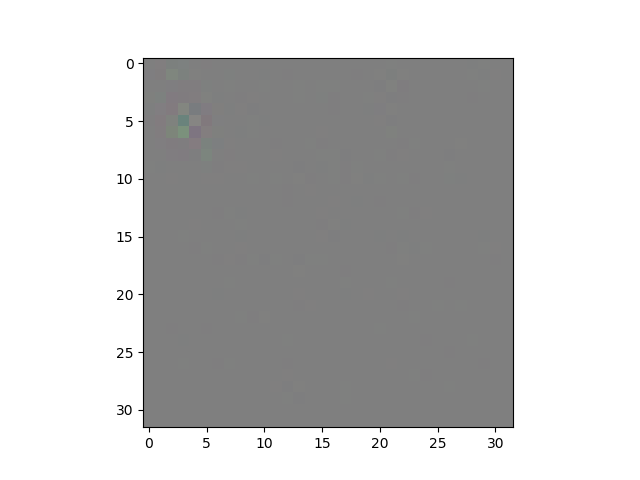}}
	\end{minipage}
	\begin{minipage}[b]{0.45\linewidth}
		\centering
		\centerline{\includegraphics[width=4cm]{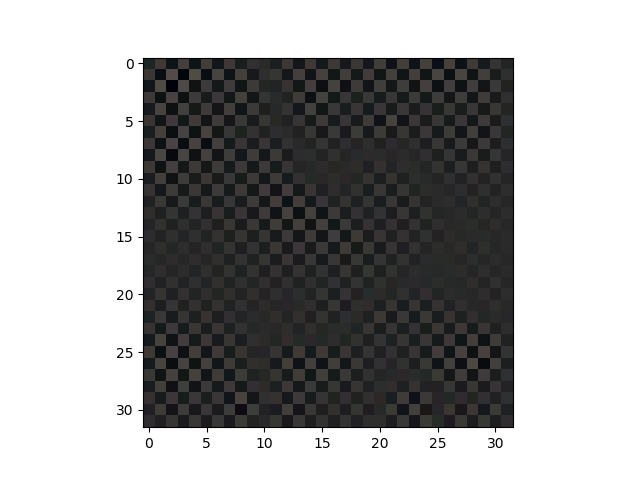}}
	\end{minipage}
	\caption{Examples of estimated backdoor patterns with ground truth a 4-pixel perturbation (left) and a global perturbation (right), respectively. The estimated global perturbation mask has been scaled by 30 times to be human visualizable.}
	\label{fig:bd_est}
	\vspace{-0.15in}
\end{figure}

In addition to accurate attack detection inference, our detection approach also gives an estimate of the ground truth backdoor pattern. The left figure of Figure \ref{fig:bd_est} is an estimate of one of the 25 4-pixel perturbations used to create the BD-P-S group. A 0.5 offset is added for visualization. The ground truth backdoor pattern is the one on the left of Figure \ref{fig:perturbation}. Comparing with the ground truth backdoor pattern, we can see that instead of perturbing all four pixels, perturbing the region near the ground truth perturbed pixel on the top left is most effective to induce group misclassification to the target class. The right figure of Figure \ref{fig:bd_est} is an estimate of one of the 25 global perturbations used to create the BD-G-S group. Since nearly half of the pixels are perturbed negatively, and the perturbation size is too small to be visualized, we first add an offset such that the resulting perturbations are all positive; then the shifted perturbation mask is scaled by 30 times. {\it Clearly, we recover a ``chess board'' pattern similar to the ground truth backdoor pattern.}

Finally, we evaluate FP using the classifiers from BD-P-S, BD-G-S and BD-G-M. We demonstrate the main weakness of FP that there is no simple ``dichotomization'' of neurons, with most {\it solely} dedicated to ``normal'' operation and some {\it solely} dedicated to implementing the backdoor. The technical details and the results are deferred to the supplementary material due to space limitations.

\subsubsection{Generalized AD approach}\label{subsubsec:bd_multi}

Our AD approach can be easily generalized, without knowledge of the attacker's mechanism, to detect imperceptible backdoors with various perturbative mechanisms. Note that most DNNs $f(\cdot)$ can be decomposed as two consecutive sub-models $f_1\circ f_2(\cdot)$, such that $f_1(\cdot):{\mathcal{X}} \rightarrow {\mathcal{Z}}$ maps an image $\underline{x}$ to its internal layer features $\underline{z}=f_1(\underline{x})\in{\mathcal Z}$, and $f_1(\cdot):{\mathcal{Z}} \rightarrow {\mathcal{C}}$ maps $\underline{z}$ to the predicted class label. Our AD framework can be generalized, taking AD-J-P as an example, by replacing Eq. (\ref{eq:obj_basic}) with the generalized objective function
\begin{equation*} \tag{10} \label{eq:obj_general}
\vspace{-0.1in}
J_{st}^{\mathcal G}(\underline{v}) = -\frac{1}{|\mathcal{D}_s|}\sum_{\underline{x}\in\mathcal{D}_s} p_{\mathcal G}(t|f_1(\underline{x})+\underline{v}),
\end{equation*}
where $p_{\mathcal G}(\cdot|\underline{z})$ is the DNN's posterior given the internal layer feature vector $\underline{z}$. That is, for each class pair $(s, t)$, we search the minimal-sized perturbation that induces high misclassification rate from $s$ to $t$ in the internal feature space ${\mathcal Z}$ instead of the image space ${\mathcal{X}}$.

By applying AD-J-P with the generalized objective function Eq. (\ref{eq:obj_general}), with $f_1(\cdot)$ the first convolutional layer, on the classifiers from the BD-P-S, BD-G-S, and BD-G-M group, we obtain detection rates 0.88, {\bf 1.00}, and {\bf 1.00}, respectively. For the clean group, we detects {\bf no attacks} for all the 25 classifiers.

We further consider a multiplicative perturbation $\underline{u}^{\ast}$ that can be applied to a clean image $\underline{x}$ by
\begin{equation*}\label{eq:bd_embedding_multi}
B_{\rm multi}(\underline{x}, \underline{u}^{\ast}) = [\underline{x} \odot \underline{u}^{\ast}]_c,
\end{equation*}
where $\underline{u}^{\ast}$ has the same dimension as the image $\underline{x}$, with $u_{ijk}$ the scaling factor of $x_{ijk}$. Here, we create a backdoor attack with a multiplicative chessboard pattern -- we set $u_{ijk}=1$ if pixel $x_{ijk}$ is not perturbed and set $u_{ijk}=1.02$ if pixel $x_{ijk}$ is perturbed. Other configurations of the attack are the same as the attacks on the classifiers in the BD-G-S group.

\begin{figure}[h]
	\begin{minipage}{1.0\linewidth}
		\vspace{-0.15in}
		\centering
		\centerline{\includegraphics[width=7cm]{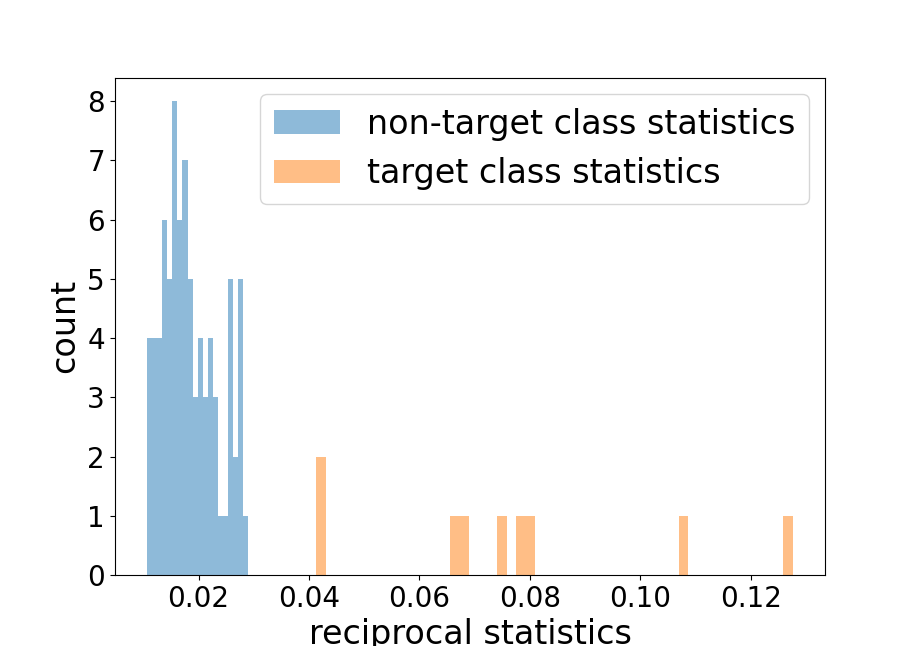}}
	\end{minipage}
	\caption{Histogram of the reciprocal statistics for AD-J-P with the generalized objective function applied on an attack with multiplicative chessboard pattern.}
	\label{fig:bd_multi_hist}
	\vspace{-0.3in}
\end{figure}

In Figure \ref{fig:bd_multi_hist}, we show the histogram of the reciprocal statistics obtained by applying AD-J-P with the generalized objective function Eq. (\ref{eq:obj_general}) on the attack above. We again choose $f_1(\cdot)$ as the first convolutional layer. The reciprocals corresponding to the true target classes are clearly, abnormally large. Our generalized AD successfully detects the attack (with order statistic p-value $2.0\times 10^{-13}\ll\theta=0.05$) with correct inference of the (source, target) class pair. In the supplementary material, we show that our generalized AD is also effective for attacks using the local patch replacement backdoor embedding mechanism assumed by NC (i.e. Eq. (\ref{eq:bd_embedding_perceptible})). While both generalized and the original additive perturbation AD approach work well for attacks with multiplicative perturbations, for attacks with the local patch replacement backdoor embedding mechanism, only the generalized AD approach is successful.

\subsection{Other Design Choices}\label{sec:OtherExp}

\subsubsection{Choice of $\pi$}\label{sec:choice_of_pi}

\begin{figure}[h]
	\vspace{-0.15in}
	\begin{minipage}{1.0\linewidth}
		\centering
		\centerline{\includegraphics[width=7.5cm]{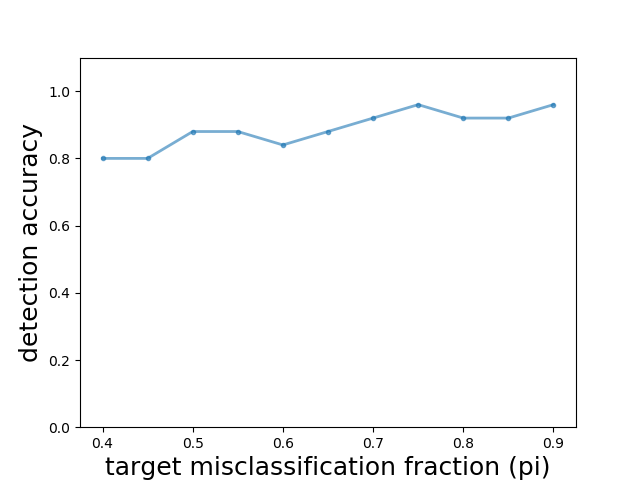}}
	\end{minipage}
	\caption{Detection accuracy of AD-J-P on the BD-P-S group, with a range of choices of $\pi$.}
	\label{fig:about_pi}
	\vspace{-0.35in}
\end{figure}

\begin{figure}[h]
	\begin{minipage}{1.0\linewidth}
		\centering
		\centerline{\includegraphics[width=7.5cm]{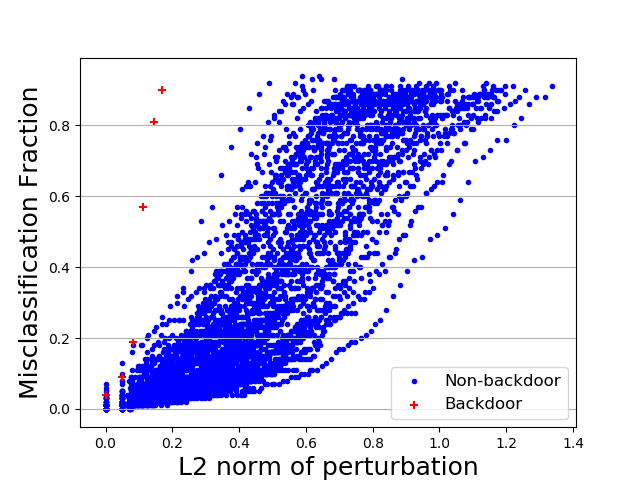}}
	\end{minipage}
	\caption{Sequences of $(||\underline{v}_{st}^{(\tau)}||_2, \rho_{st}^{(\tau)})$ for all $(s, t)$ pairs, including the ground truth backdoor pair $(s^{\ast}, t^{\ast})$ represented using red crosses, during the execution of Algorithm \ref{alg:main} with $\pi=0.9$, while applying AD-J-P on an example classifier realization in the BD-P-S group.}
	\label{fig:about_pi2}
	\vspace{-0.15in}
\end{figure}

Unlike NC, the proposed detection approach is not very sensitive to the choice of $\pi$. As an example, we evaluate the detection accuracy of the AD-J-P variant on the 25 classifiers in the BD-P-S group, with a range of choices of $\pi$ from 0.4 to 0.9. We use the default detection threshold $\theta=0.05$. As shown in Figure \ref{fig:about_pi}, the detection accuracy does increase, but not dramatically with $\pi$. The minimum detection accuracy across the range of choices of $\pi$ is 0.8, which can be further improved by use of a more liberal detection threshold. This phenomenon can be understood from Figure \ref{fig:about_pi2}, in which we plot the sequences of $(||\underline{v}_{st}^{(\tau)}||_2, \rho_{st}^{(\tau)})$ for all $(s, t)$ pairs during the perturbation optimization using Algorithm \ref{alg:main} with $\pi=0.9$, while applying AD-J-P on an example classifier realization in the BD-P-S group. The sequence corresponding to the ground truth backdoor class pair $(s^{\ast}, t^{\ast})$ is represented using red crosses, and is clearly separated from the sequences for the non-backdoor pairs. In other words, there is a huge range for the choice of $\pi$ (not exceeding 0.9) to achieve correct detection for this example. For any choice of $\pi$ in such range, the $(||\underline{v}_{st}^{(\tau)}||_2, \rho_{st}^{(\tau)})$ sequence for each $(s, t)$ pair is truncated at the minimum $\tau$ such that $\rho_{st}^{(\tau)}$ first exceeds $\pi$. The resulting $||\underline{v}_{st}^{(\tau)}||_2$ for the backdoor class pair $(s^{\ast}, t^{\ast})$, as can be seen from the figure, will be much smaller than those for the non-backdoor pairs, which will lead to a successful detection.

\subsubsection{Size of Clean Data Set}

\begin{figure}[h]
	\begin{minipage}{1.0\linewidth}
		\centering
		\centerline{\includegraphics[width=7cm]{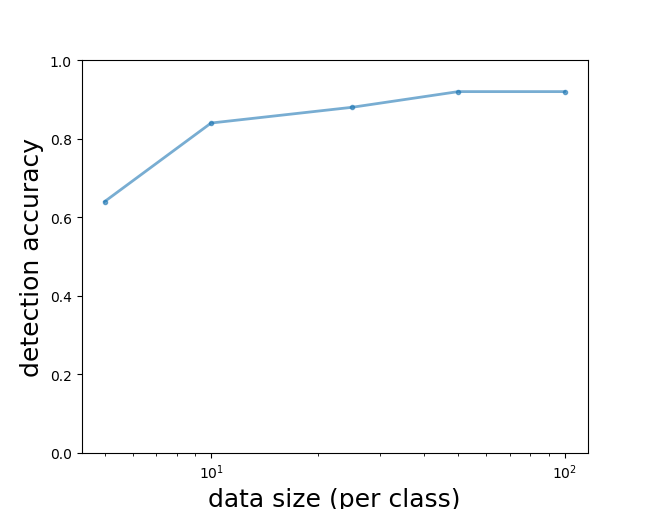}}
	\end{minipage}
	\caption{Detection accuracy of AD-J-P on BD-P-S for a range of sizes of the clean data set.}
	\label{fig:about_Ni}
	\vspace{-0.15in}
\end{figure}

Here we test the impact on detection accuracy of the size of the clean data set available to the defender. Again, we apply AD-J-P detection (with $\theta=0.05$) to the classifiers in the BD-P-S group, but with the number of clean images per class used for detection equaling 5, 10, 25, 50, 100 for each trial, respectively. As shown in Figure \ref{fig:about_Ni}, the detection accuracy increases as the number of clean images per class grows, approaching a stably high value when the number of clean images per class is greater than 25. Note that detection accuracy greater than $80\%$ is achieved with just 10 clean samples per class. Comparing with the size of the training set (5000 images per class) and the number of backdoor images required to launch the attack (1000 images in total\footnote{For backdoor patterns with innocuous objects, poisoning the training set with a few to a few tens of backdoor images yields a successful attack \cite{Song}. For backdoor patterns with human-imperceptible perturbations, hundreds of backdoor images should be used for poisoning the training set to make the attack successful \cite{SS}.}), the detection cost is ``cheap'' in terms of the required number of clean images. Even if there are abundant clean images available to the defender, he/she can always use only a subset of images to reduce the defender's computational effort.

\subsubsection{Fitting Null Distribution Using All Statistics}\label{sec:all_stats}

As described in Section \ref{sec:inference}, during the detection inference process, we fit a conditional null density using the $(K-1)^2$ smallest statistics. One may also consider a naive approach that fits an unconditional null using all detection statistics. In Table \ref{tab:naive}, we show the detection accuracy of the AD-J-P variant with detection threshold $\theta=0.05$ for the four groups of DNN classifiers based on this naive null learning. Clearly, there is severe degradation in accuracy of detecting backdoors.

\begin{table}[h]
	\begin{center}
		\caption{Detection accuracy of the AD-J-P variant with detection threshold $\theta=0.05$ for the four groups of DNN classifiers, where the null density is estimated using all $K(K-1)$ reciprocal statistics.}
		\resizebox{0.45\textwidth}{!}{
			\begin{tabular}{ |c|c|c|c|c| } 
				\hline
				& BD-P-S & BD-G-S & BD-G-M & Clean \\ 
				\hline
				detection acc. & 0.48 & 0.44 & 0.52 & {\bf 1.00}\\
				\hline
			\end{tabular}\label{tab:naive}}
	\end{center}
	\vspace{-0.15in}
\end{table}

\subsubsection{Size/Energy of the Ground Truth Backdoor Pattern}\label{sec:size_of_gt_bd}

Here we create 25 classifier realizations using the same setting as for the BD-G-S group, except that the L2 norm of the backdoor pattern is set to 1.2 -- much larger than required to launch a successful backdoor attack (and visualizable by carefully scrutinizing the image). When applying AD-J-P for detection, with the detection threshold $\theta=0.05$, the detection accuracy reaches $1.00$. More importantly, the estimated backdoor patterns for the 25 classifier realizations are all scaled ``chess board'' patterns with L2 norm around $0.101\pm0.017$, while the L2 norm of the optimized perturbation for all non-backdoor class pairs across the 25 realizations is $1.243\pm0.321$. That is, the ``estimated'' backdoor pattern will have the {\it minimum} norm necessary to achieve the $\pi$ target level, even if the actual backdoor pattern has much larger norm. Regardless, the estimated backdoor pattern {\it is} accurate (it is a ``chess board'' pattern). Also, from this experiment, we note that the proposed detector does not require the size/energy of the ground truth backdoor pattern to be significantly smaller than the size/energy of perturbations required to induce group misclassification for non-backdoor class pairs.
\vspace{-0.1in}
\subsection{Computational Complexity}

The computation in this work is performed on an Amazon EC2 g3s.xlarge virtual machines powered by a NVIDIA Tesla M60 GPU. When applying AD-J-P to the BD-P-S, the average running time across the 25 classifier realizations is 698s. In comparison, when applying NC-L1-1.5 to BD-P-S, the average running time across the 25 classifier realizations is 912s. For both approaches, the perturbation optimization step incurs most of the computational cost. For the NC defense, each iteration of perturbation optimization requires $2(K-1)N$ back-propagations on the DNN, where $N$ is the number of clean images per class used for detection. Since NC solves $K$ perturbation optimization problems, the total computational cost is $2(K-1)NTK$ back-propagations, where $T$ is the number of iterations required to reach $\pi$-level misclassification. For our AD framework, each iteration of perturbation optimization requires $2N$ back-propagations on the DNN since we only use clean images from a single class for detection. The total computational cost is then $2NTK(K-1)$ since we solve $K(K-1)$ perturbation optimization problems. Theoretically, the computational cost of the two defenses are the same. But in practice, the difficulty of inducing group misclassification to a target class from all the other classes can be much higher than inducing misclassification from a single source class to the target class, especially when the backdoor attack is crafted with a single source class. Therefore, the number of iterations in perturbation optimization required for the NC defense is generally larger than that for our proposed defense.

\vspace{-0.1in}
\section{Conclusions and Future Work}

In this work, we developed a purely unsupervised AD defense that detects imperceptible backdoor attacks in DNN classifiers post-training. We tested multiple variants of our AD framework and compared them with other existing post-training defenses for several backdoor patterns, data sets (shown in the supplementary material), and attack settings. Our defense was experimentally shown to be more effective and robust over a range of detection scenarios. 

There are several directly related issues that have been considered and can be further explored. First, although we proposed a correction scheme accounting for class pairs with high confusion, it is possible for two classes to be very similar but for their confusion to be low. In such case, the minimum perturbation size required to induce group misclassification for such class pairs may be as small as for a true backdoor pair. We believe it is possible to devise an extension of our scheme that can distinguish such false pairs from true ones. Second, our proposed defense focuses on detecting backdoor attacks with human-imperceptible backdoor patterns. If the backdoor pattern is an innocuous object in the image
({\it e.g.}  a tennis ball in a dog image or glasses on a face), the norm of the backdoor pattern might be much larger than the average norm of the minimum perturbation required to induce group misclassification for non-backdoor class pairs. Such backdoors violate our fundamental assumption about the attacker and require an alternative detection strategy. In fact, we have recently devised a quite effective approach for this case \cite{Perceptible_MLSP}.
Third, applying perturbation optimization at an internal layer, in addition to addressing more general backdoor attack mechanisms, also address other domains -- particularly where the inputs to the DNN are {\it discrete-valued} ({\it e.g.}, text) and, thus, for which gradient-based optimization of a (continuous-valued) perturbation cannot be applied.  Fourth, backdoor attacks and defenses can be considered for classifiers other than DNNs, {\it e.g.} support vector machines. 

There are several other related topics that can be further explored. First, while our cost function minimization was proposed as part of a backdoor attack detection scheme, it can also in principle be applied to provide {\it interpretability} to existing (unattacked) DNN classifier solutions. In particular, our approach may identify a class pair $(s,t)$ that is highly confusable in the presence of an unusually small perturbation.  This is revealing of a possible hidden fragility of the learned classifier.  Moreover, it is possible that retraining with an augmented training set that includes perturbed examples from class $s$ labeled as class $s$ may help to remedy such fragilities. Second, a defense ``in-flight'' can be developed to identify backdoor patterns during the operation of the classifier. For example, the estimated backdoor pattern revealed by our approach may be used to correlate against test patterns. Third, we can develop defenses against backdoor attacks while assuming a much stealthier attacker. For example, the attacker may be able to optimize the backdoor pattern, to make the backdoor less detectable. Or we may study defenses against ``white box'' attacks where the attack is assumed to be aware of the deployed defense. Moreover, a more surgical attack on $(s^{\ast}, t^{\ast})$ may be devised by inserting images crafted using clean images from all classes other than $s^{\ast}$ or $t^{\ast}$, added with the backdoor pattern and {\it correctly} labeled, into the training set together with backdoor images to minimize ``collateral damage''.

\vspace{-0.1in}
\bibliographystyle{plain}
\bibliography{Biblio}

\end{document}